\documentclass{article}

\usepackage{arxiv}

\usepackage[utf8]{inputenc} % allow utf-8 input
\usepackage[T1]{fontenc}    % use 8-bit T1 fonts
\usepackage{hyperref}       % hyperlinks
\usepackage{url}            % simple URL typesetting
\usepackage{booktabs}       % professional-quality tables
\usepackage{amsfonts}       % blackboard math symbols
\usepackage{nicefrac}       % compact symbols for 1/2, etc.
\usepackage{microtype}      % microtypography
\usepackage{lipsum}
\usepackage{graphicx}
\usepackage{subfig}
\usepackage{amsmath}
\usepackage{multirow}
\usepackage{authblk}
\graphicspath{ {./images/} }
\newtheorem{definition}{Definition}

\title{A Review of Deep Learning Methods for Irregularly Sampled Medical Time Series Data}

\author[1,2]{Chenxi Sun}
\author[3,4]{Shenda Hong}
\author[1,2]{Moxian Song}
\author[1,2]{Hongyan Li \thanks{Corresponding author: leehy@pku.edu.cn}}
\affil[1]{Key Laboratory of Machine Perception (Ministry of Education), Peking University, Beijing, China.}
\affil[2]{School of Electronics Engineering and Computer Science, Peking University, Beijing, China.}
\affil[3]{National Institute of Health Data Science, Peking University, Beijing, China.}
\affil[4]{Institute of Medical Technology, Health Science Center, Peking University, Beijing, China.}

\begin{document}
\maketitle

\begin{abstract}
Irregularly sampled time series (ISTS) data has irregular temporal intervals between observations and different sampling rates between sequences. ISTS commonly appears in healthcare, economics, and geoscience. Especially in the medical environment, the widely used Electronic Health Records (EHRs) have abundant typical irregularly sampled medical time series (ISMTS) data. Developing deep learning methods on EHRs data is critical for personalized treatment, precise diagnosis and medical management. However, it is challenging to directly use deep learning models for ISMTS data. On the one hand, ISMTS data has the intra-series and inter-series relations. Both the local and global structures should be considered. On the other hand, methods should consider the trade-off between task accuracy and model complexity 
and remain generality and interpretability. So far, many existing works have tried to solve the above problems and have achieved good results. In this paper, we review these deep learning methods from the perspectives of technology and task. Under the technology-driven perspective, we summarize them into two categories - missing data-based methods and raw data-based methods. Under the task-driven perspective, we also summarize them into two categories - data imputation-oriented and downstream task-oriented. For each of them, we point out their advantages and disadvantages. Moreover, we implement some representative methods and compare them on four medical datasets with two tasks. Finally, we discuss the challenges and opportunities in this area.
\end{abstract}

\section{Introduction}
\label{sec:introduction}

Time series data have been widely used in practical applications, such as health \cite{1}, geoscience \cite{2}, sales \cite{3}, and traffic [4]. The popularity of time series prediction, classification, and representation has attracted increasing attention, and many efforts have been taken to address the problem in the past few years \cite{8,9,10,11}.

The majority of the models assume that the time series data is even and complete. However, in the real world, the time series observations usually have non-uniform time intervals between successive measurements. Three reasons can cause this characteristic: 1) The missing data exists in time series due to broken sensors, failed data transmissions or damaged storage. 2) The sampling machine itself does not have a constant sampling rate. 3) Different time series usually comes from different sources that have various sampling rates. We call such data as irregularly sampled time series (ISTS) data. ISTS data naturally occurs in many real-world domains, such as weather/climate \cite{2}, traffic \cite{4}, and economics \cite{3}. 

In the medical environment, irregularly sampled medical time series (ISMTS) is abundant. The widely used electronic health records (EHRs) data have a large number of ISMTS data. EHRs are the real-time, patient-centered digital version of patients’ paper charts. EHRs can provide more opportunities to develop advanced deep learning methods to improve healthcare services and save more lives by assisting clinicians with diagnosis, prognosis, and treatment \cite{36}. Many works based on EHRs data have achieved good results, such as mortality risk prediction \cite{31,52}, disease prediction \cite{37,38,39}, concept representation \cite{41,42} and patient typing \cite{29,42,43}.

Due to the special characteristics of ISMTS, the most important step is establishing the suitable models for it. However, it is especially challenging in medical settings. 

Various tasks need different adaptation methods. Data imputation and prediction are two main tasks. The data imputation task is a processing task when modeling data, while the prediction task is a downstream task for the final goal. The two types of tasks may be intertwined. Standard techniques, such as mean imputation \cite{93}, singular value decomposition (SVD) \cite{94} and k-nearest neighbour (kNN) \cite{95}, can impute data. But they still lead to the big gap between the calculated data distribution and have no ability for the downstream task, like mortality prediction. Linear regression (LR) \cite{96}, random forest (RF)\cite{97} and support vector machines (SVM) \cite{98} can predict, but fails for ISTS data.

State-of-the-art deep learning architectures have been developed to perform not only supervised tasks but also unsupervised ones that relate to both imputation and prediction tasks. Recurrent neural networks (RNNs) \cite{100,101,102}, auto-encoder (AE) \cite{103,104} and generative adversarial networks (GANs) \cite{105,106} have achieved good performance in medical data imputation and medical prediction thanks to their abilities of learning and generalization obtained by complex nonlinearity. They can carry out prediction task or imputation task separately, or they can carry out two tasks at the same time through the splicing of neural network structure.

\begin{table*}[t]
\caption{Abbreviations}\label{tb:abbreviations}
\centering
\begin{tabular}{|l|l|l|l|l|l|}
\hline
Full name  &Abbreviations &Full name  &Abbreviations\\
\hline
Time series &TS &Recurrent  neural  network &RNN\\
\hline
Irregularly sampled time series &ISTS &Long short-term unit &LSTM\\
\hline
Irregularly sampled medical time series &ISMTS &Gated  recurrent  unit &GRU\\
\hline
Electronic health record &EHR &Generative adversarial  network &GAN\\
\hline
\end{tabular}
\end{table*}

Different understandings about the characteristics of ISMTS data appear in existing deep learning methods. We summarized them as missing data-based perspective and raw data-based perspective. The first perspective \cite{1,24,64,27,62} treat irregular series as having missing data. They solve the problem through more accurate data calculation. The second perspective \cite{29,54,63,55,56} is on the structure of raw data itself. They model ISMTS directly through the utilization of irregular time information. Neither views can defeat the other. 

Either way, it is necessary to grasp the data relations comprehensively for more effectively modeling. We conclude two relations of ISMTS - intra-series relations (data relations within a time series) and inter-series relations (data relations between different time series). All the existing works model one or both of them. They relate to the local structures and global structures of data and we will introduced in Section \ref{sec:characteristics}.

Besides, different EHR datasets may lead to different performance of the same method. For example, the real-world MIMIC-III \cite{32} and CINC\cite{91} datasets record multiple different diseases. The records between diseases have distinct data characteristics and the prediction results of each general methods \cite{1,29,64,62} varied between each disease datasets. Thus, many existing methods model a specific disease record, like sepsis \cite{107}, atrial fibrillation\cite{108,109} and kidney disease\cite{110} and have improved the predicting accuracy.

The rest of the paper is organized as follows. Section \ref{sec:preliminaries} gives the basic definition and abbreviations. Section \ref{sec:characteristics} describes the features of ISMTS based on two viewpoints - intra-series and inter-series. Section \ref{sec:categories1} and Section \ref{sec:categories2} introduce the related works in technology-driven perspective and task-driven perspective. In each perspective, we summarize the methods into specific categories and analyze the merits and demerits. Section \ref{sec:experiments} compares the experiments of some methods on four medical datasets with two tasks. In section \ref{sec:discussions} and \ref{sec:conclusion}, we raise the challenges and opportunities for modeling ISMTS data and then make conclusion.

\begin{figure*}[t]
\centerline{\includegraphics[width=0.7\linewidth]{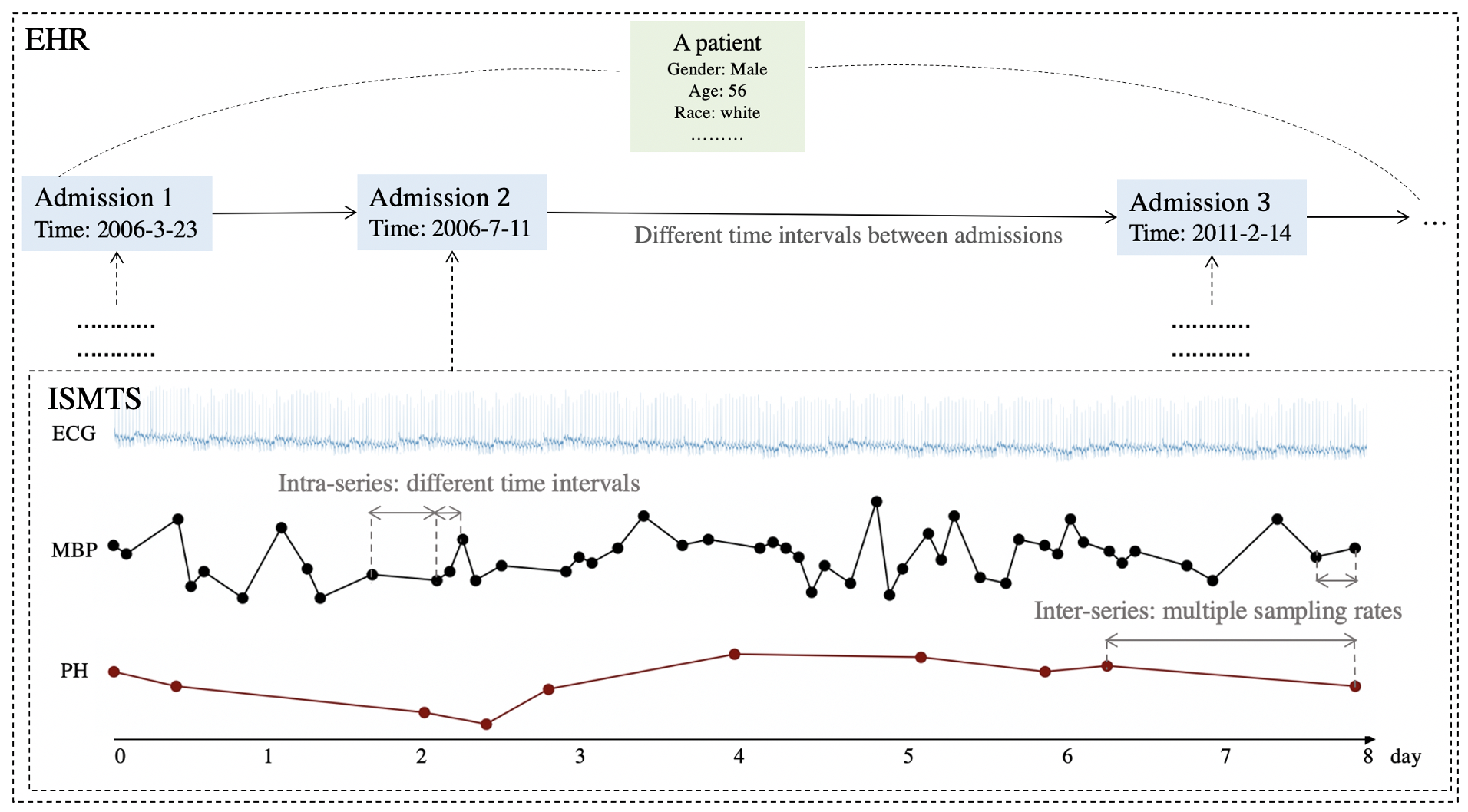}}
\caption{An example of a patient’s temporal EHR}
\label{fig:EHRs}
\end{figure*}

\section{Preliminaries}\label{sec:preliminaries}

The summary of abbreviations is in Table \ref{tb:abbreviations}.

A typical EHR dataset is consist of a number of patient information which includes demographic information and in-hospital information. In-hospital information is a hierarchical patient-admission-code form shown in Figure \ref{fig:EHRs}. Each patient has certain admission records as he/she could be in hospital several times. The codes have diagnoses, lab values and vital sign measurements. 

\begin{definition}[Electronic Health Records data EHRs] \label{def:EHRs}
An electronic health record consists a set of patient records $\mathcal{R}=\{r_i|i=0,...,N-1\}$, where $N$ is the number of records. A patient $p_i$, having the demographic information $I_i$, may have $K_i$ admission records. $R_i$ is the admission record set of a patient $p_i$, $R_i=\{r_j\in R|j=0,...,K_{i}-1\}$. Each record $r_i$ is consist of many codes, including static diagnoses codes set ${d_i}$ and dynamic vital signs codes set $x_i$. Each code has the time stamp $t$.
\end{definition}

EHRs have many ISMTS because of two aspects: 1) multiple admissions of one patient and 2) multiple time series records in one admission. Multiple admission records of each patient have different time stamps. Because of health status dynamics and some unpredictable reasons, a patient will visit hospitals under varying intervals \cite{46}. For example, in Figure 1, March 23, 2006, July 11, 2006 and February 14, 2011 are patient admission times. The time interval between the 1st admission and 2nd admission is couple of months while the time interval between admissions 2, 3 is 5 years. Each time series, like blood pressure in one admission, also has different time intervals. Shown as Admission 2 in Figure \ref{fig:EHRs}, the sampling time is not fixed. Different physiological variables are examined at different times due to the changes in symptoms. Every possible test is not regularly measured during an admission. When a certain symptom worsens, corresponding variables are examined more frequently; when the symptom disappears, the corresponding variables are no longer examined. 

Without the loss of generality, we only discuss univariate time series. Multivariate time series can be modeled in the same way.

\begin{definition}[\small Irregularly Sampled Medical Time Series ISMTS] \label{def:ISMTS}
A dataset of ISMTS $\mathcal{D}=\{(s_i,y_i)\in(S,Y)|i=0,...,N-1\}$ has $N$ time series. $s_i$ is the observation of $i$th simple and $y_i$ is its label. $s_i$ is represented as a tuple $(x_i,t_i )=([x_{i,0},...,x_{i,L_{i}-1}],[t_{i,0},...,t_{i,L_{i}-1}])$ with $L_i$ time steps. $x_{i,l}$ is the observed value in time step l and $t_{i,l}$ is the corresponding time. The time interval between the $p$th time step and the $q$th time step of $s_i$ is $\delta_{i,(p,q)}$. Time interval $\delta$ varies between different neighboring time steps. 
\end{definition}

\begin{equation}
\delta_{i,(p,q)}=|t_{i,p}-t_{i,q}|
\end{equation}

Definition \ref{def:ISMTS} illustrates three important matters of ISMTS - the value $x$, the time $t$ and the time interval $\delta$. In some missing value-based works (we will introduce in Section \ref{sec:categories1}), they use masking vector $m\in\{0,1\}$ to represent the missing value. 

\begin{equation}
m_{i}=
\left\{ \begin{aligned} 
&1  \quad\quad  \textit{if $x_{i}$ is observed}\\
&0  \quad\quad  otherwise
\end{aligned} \right.
\end{equation}

\section{Characteristics of irregularly sampled medical time series}
\label{sec:characteristics}

The medical measurements are frequently correlated both within streams and across streams. For example, the value of blood pressure of a patient at a given time could be correlated with the blood pressure at other times, and it could also have a relation with the heart rate at that given time. Thus, we will introduce ISMTS's irregularity in two aspects: 1) intra-series and 2) inter-series.

Intra-series irregularity is the irregular time intervals between the nearing observations within a stream. For example, shown in Figure \ref{fig:EHRs}, the blood pressure time series have different time intervals, such as 1 hour, 2 hours, and even 24 hours. The time intervals add a time sparsity factor when the intervals between observations are large \cite{46}. Existing two ways can handle the irregular time intervals problem: 1) Determining a fixed interval, treating the time points without data as missing data. 2) Directly modeling time series, seeing the irregular time intervals as information. The first way requires a function to impute missing data \cite{48,49}. For example, some RNNs \cite{1,26,27,28,51,53} can impute the sequence data effectively by considering the order dependency. The second way usually uses the irregular time intervals as inputs. For example, some RNNs \cite{29,54} apply time decay to affect the order dependency, which can weaken the relation between neighbors with long time intervals.

Inter-series irregularity is mainly reflected in the multi-sampling rates among different time series. For example, shown in Figure \ref{fig:EHRs}, vital signs such as heart rate (ECG data) have a high sampling rate (in seconds), while lab results such as pH (PH data) are measured infrequently (in days) \cite{44,45}. Existing two ways can handle the multi-sampling rates problem: 1) Considering data as a multivariate time series. 2) Processing multiple univariable time series separately. The first way aligns the variables of different series in the same dimension and then solves the missing data problem \cite{47}. The second way models different time series simultaneously and then designs fusion methods \cite{55}.

Numerous related works are capable of modeling ISMTS data, we category them from two perspectives: 1) technology-driven and 2) task-driven. We will describe each category in detail.

\begin{figure*}[t]
\centering{
\includegraphics[width=0.6\linewidth]{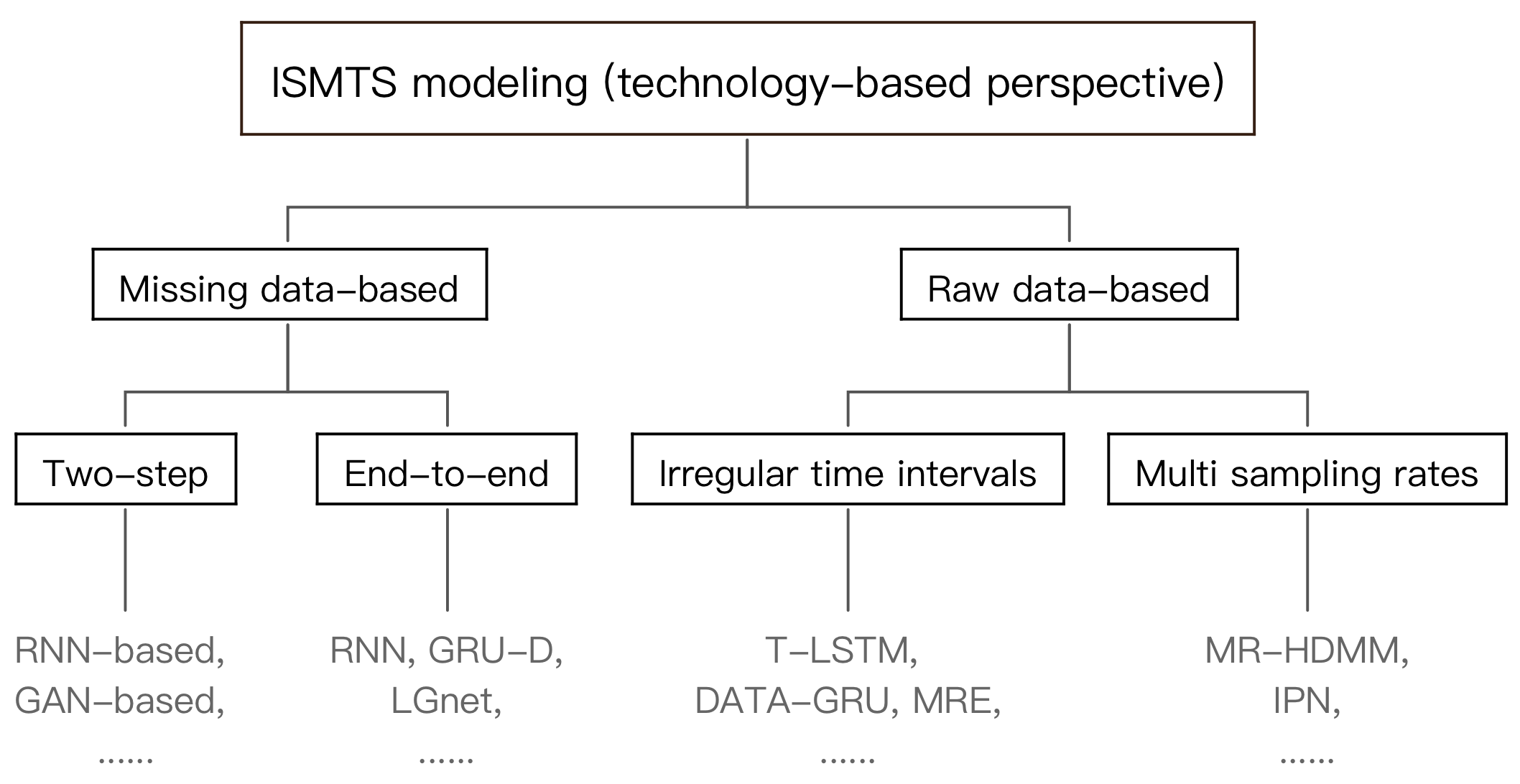}
}
\caption{Categories based on technology-driven}
\label{fig:tech_driven}
\end{figure*}

\begin{figure*}[t]
\centering
\subfloat[Missing rates of Physionet dataset]{
\includegraphics[width=0.25\linewidth]{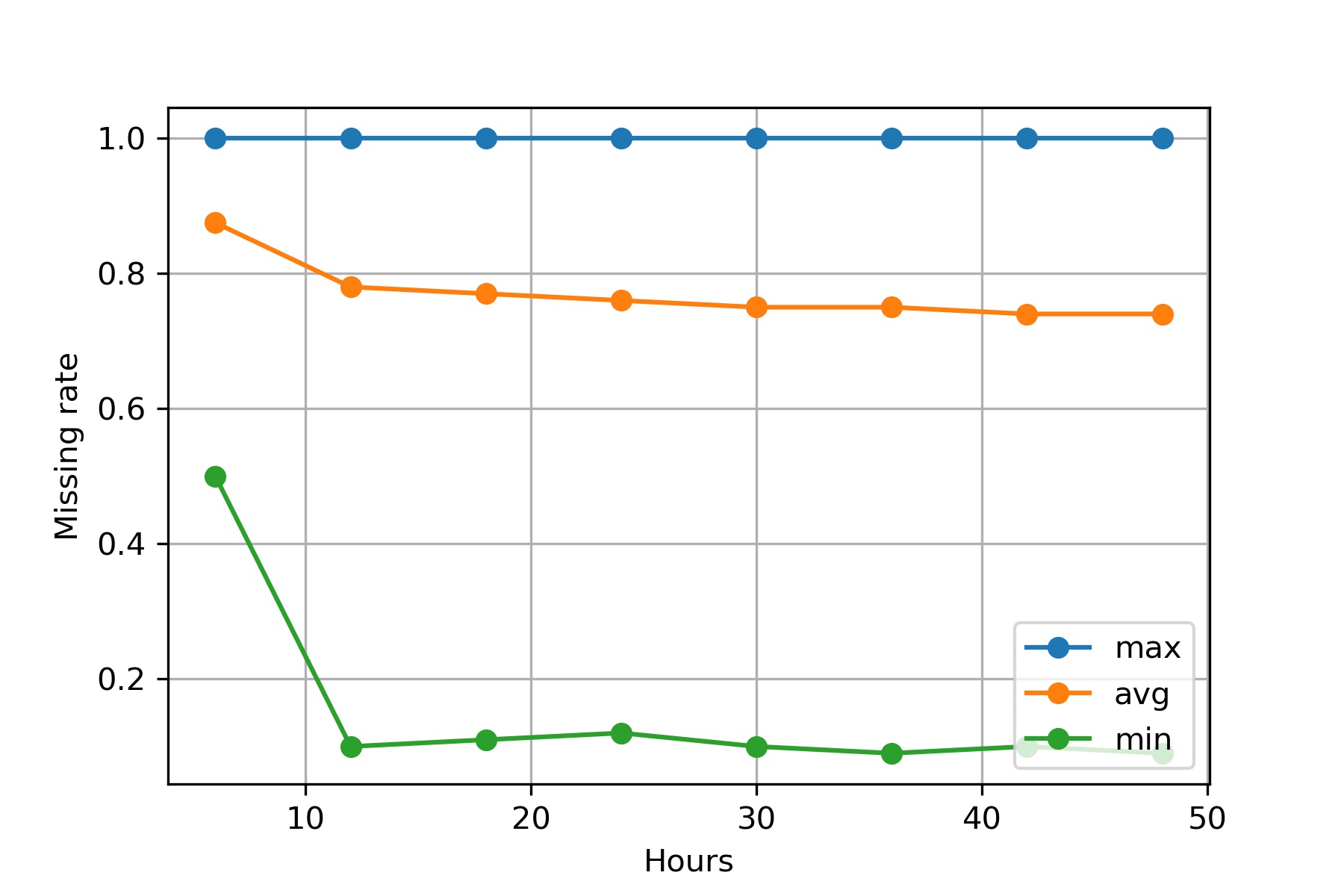}
\label{fig:pysionet_missing}
}
\subfloat[Missing rates of of MIMIC-III dataset]{
\includegraphics[width=0.25\linewidth]{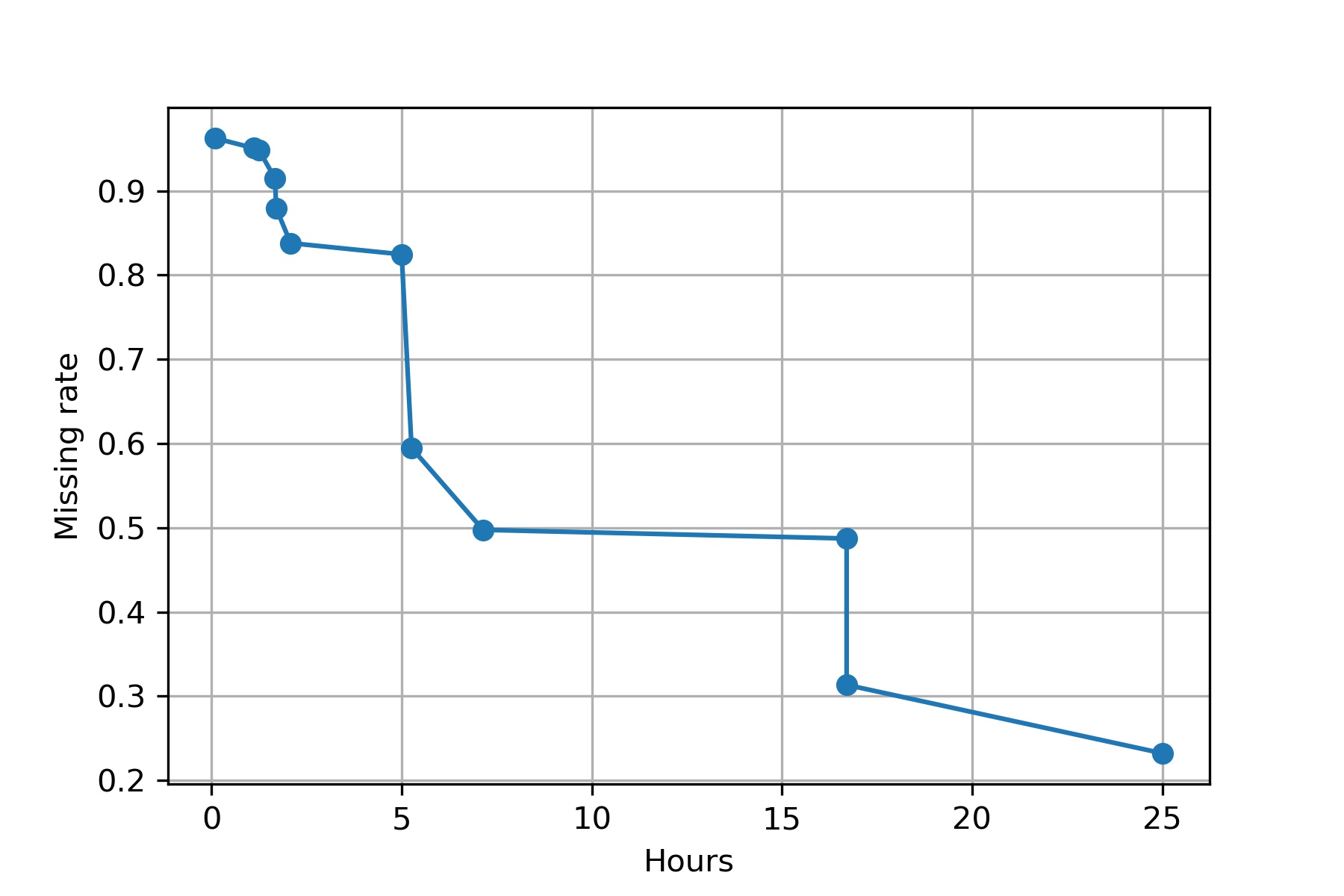}
\label{fig:mimic_missing}
}
\subfloat[Missing rates of sepsis dataset]{
\includegraphics[width=0.25\linewidth]{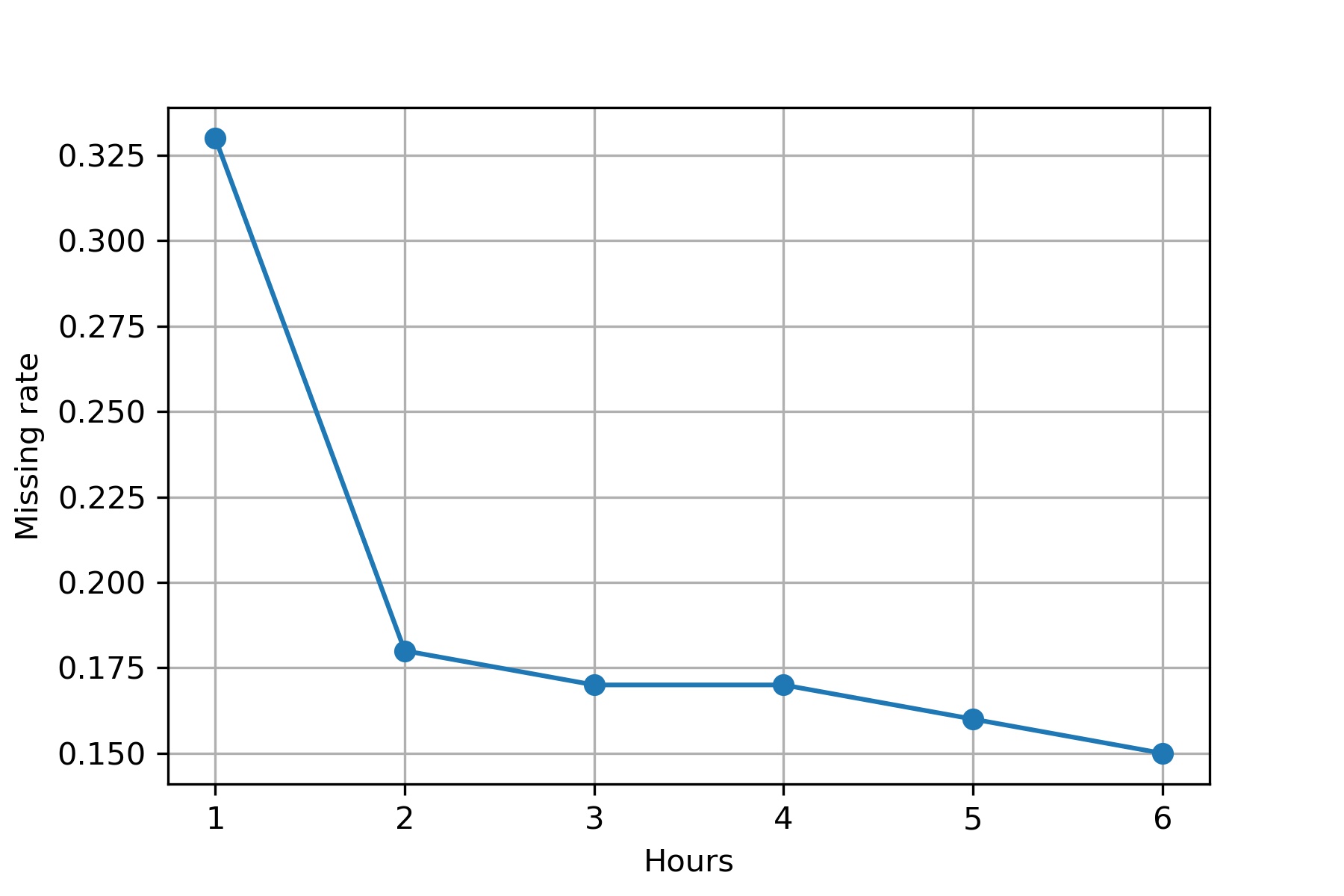}
\label{fig:sepsis_missing}
}
\subfloat[Missing rates of COVID-19 dataset]{
\includegraphics[width=0.25\linewidth]{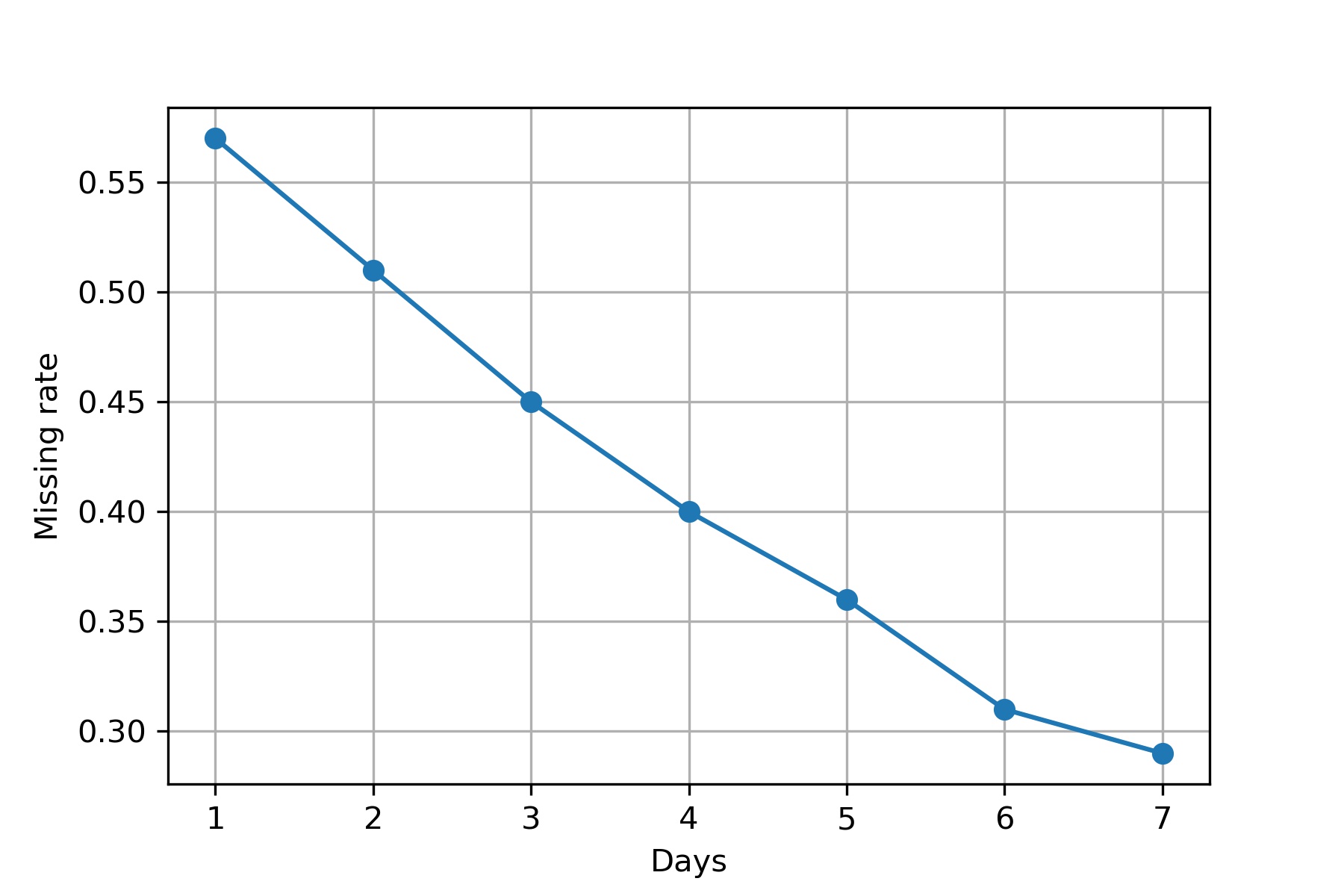}
\label{fig:covid_missing}
}
\caption{The missing rates of real-world EHRs data. \\
\small
(a) The lines stand for maximum, minimum and average $r_{missing}$ at each hour. The global $r_{missing}=80.67\%$.\\
(b) The global $r_{missing}=85.02\%$. $r_{missing}>90\%$ when the $r_{sampling}=1$ (per hour).\\
(c) The global $r_{missing}=23.42\%$. $r_{missing}$ drops sharply when $r_{sampling}=0.5$ (per hour)\\
(d) The global $r_{missing}=85.00\%$. $r_{missing}<50\%$ until $r_{sampling}<1$ (per day).
}
\label{fig:missing}
\end{figure*}

\section{Categorization based on technology-driven} \label{sec:categories1}

Based on technology-driven, we divide the existing works into two categories: 1) missing data-based perspective and 2) an raw data-based perspective. The specific categories are shown in Figure \ref{fig:tech_driven}.

The missing data-based perspective regards every time series has uniform time intervals. The time points without data are considered to be the missing data points. As shown in Figure \ref{fig:align}, when converting irregular time intervals to regular time intervals, missing data shows up. The missing rate $r_{missing}$ can measure the degree of the missing at a given sampling rate $r_{sampling}$. 

\begin{equation}
r_{missing}=\frac{\textit{\# of time points with missing data}}{\textit{\# of time points}}
\end{equation}

\begin{equation}
r_{sampling}=\frac{\textit{\# of observed data}}{\textit{time unit}}
\end{equation}

\noindent The ISMTS in the real-world EHRs have a severe problem with missing data. For example, Luo et al. \cite{13} gathered statistics of CINC2012 dataset \cite{31,92}. As time goes by, the results show that the maximum missing rate at each timestamp is always higher than 95\%. Most variables’ missing rate is above 80\%, and the mean of the missing rate is 80.67\%, as shown in Figure \ref{fig:pysionet_missing}. The other three real-word EHRs data set MIMIC-III dataset \cite{32},CINC2019 dataset \cite{58,91}, and COVID-19 dataset \cite{57} are also affected by the missing data, shown in Figure \ref{fig:mimic_missing}, \ref{fig:sepsis_missing}, and \ref{fig:covid_missing}. In this viewpoint, existing methods impute the missing data, or model the missing data information directly. 

The raw data-based perspective uses irregular data directly. The methods do not fill in missing data to make the irregular sampling regular. On the contrary, they think that irregular time itself is the valuable information. As shown in Figure \ref{fig:irregular_intervals}, the time are still irregular and the time intervals are recorded. Irregular time intervals and multi-sampling rates are intra-series characteristic and inter-series characteristic we have introduced in Section \ref{sec:characteristics} respectively. They are very common phenomenons in EHR database. For example, CINC2019 dataset is relatively clean but still has more than 60\% samples with irregular time intervals. Only 1.28\% samples have the same sampling rate in MIMIC-III dataset. In this viewpoint, methods usually integrate the features of varied time intervals to the inputs of model, or design models which can process samples with different sampling rates.

\begin{figure*}[t]
\centerline{\includegraphics[width=0.55\linewidth]{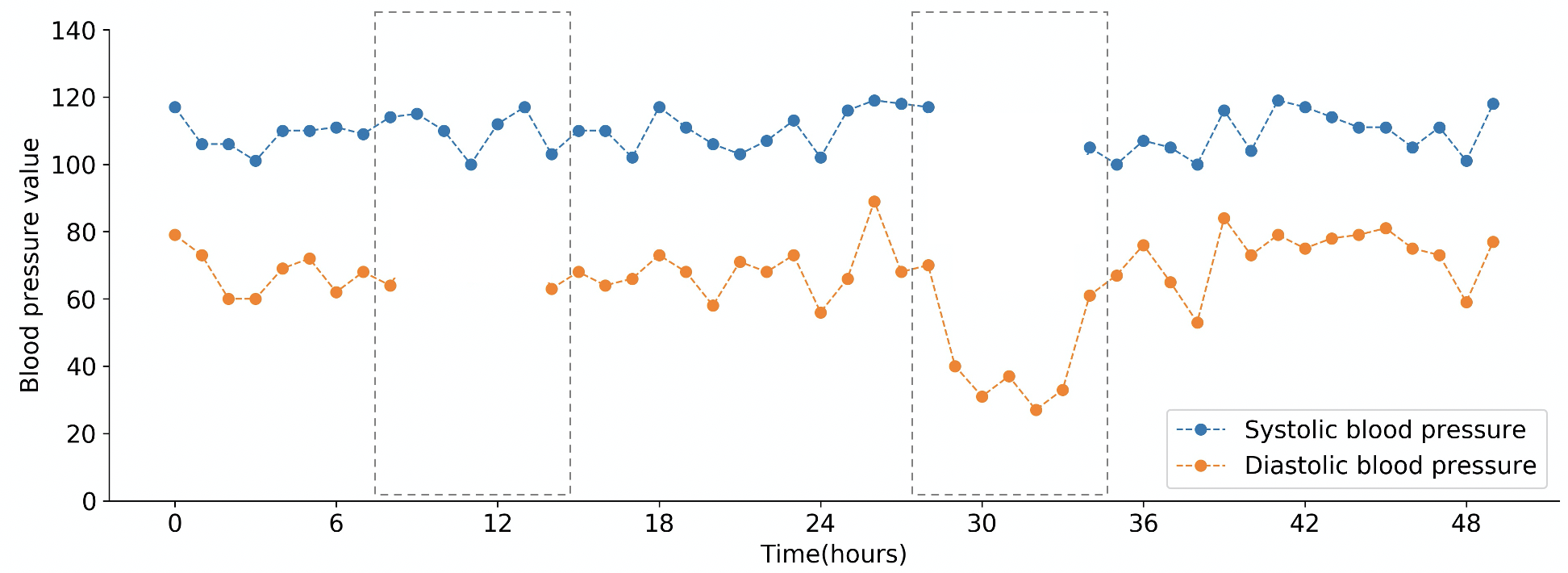}}
\caption{Systolic and diastolic blood time series with missing data of a patient in ICU}
\label{fig:important_missing}
\end{figure*}

\begin{figure*}[t]
\centering
\subfloat[Convert irregularly sampled data to missing data – an example of the pH value of a patient]{
\includegraphics[width=0.8\linewidth]{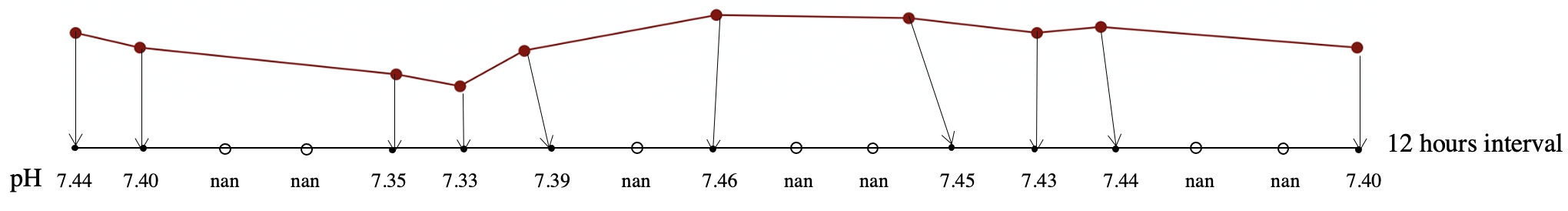}
\label{fig:align}
}\\
\subfloat[Record the irregular time intervals between observations – an example of the pH value of a patient]{
\includegraphics[width=0.8\linewidth]{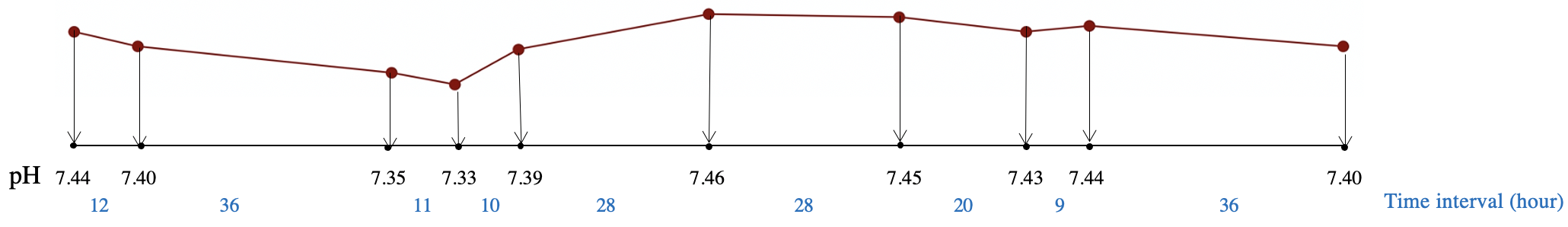}
\label{fig:irregular_intervals}
}
\caption{Two perspectives of modeling irregularly sampled medical time series}
\label{fig:aligns}
\end{figure*}

\subsection{Missing data-based perspective} \label{sec:missing}

The methods of missing data-based perspective convert ISMTS into equally spaced data. They \cite{72,73,74} discretize the time axis into non-overlapping intervals with hand-designed intervals. Then the missing data shows up. 

The missing values damage temporal dependencies of sequences \cite{13} and make applying many existing models directly infeasible, such as linear regression \cite{6} and recurrent neural networks (RNN) \cite{7}. As shown in Figure \ref{fig:important_missing}, because of missing values, the second valley of the blue signal is not observed and cannot be inferred by simply relying on existing basic models \cite{6,7}. But the valley values of blood pressure are significant for ICU patients to indicate sepsis \cite{15}, a leading cause of patient mortality in ICU \cite{14}. Thus, missing values have an enormous impact on data quality, resulting in unstable predictions and other unpredictable effects \cite{16}. Many prior efforts have been dedicated to the models that can handle missing values in time series. And they can be divided into two categories: 1) two-step approaches and 2) end-to-end approaches.

\begin{figure*}[t]
\centerline{\includegraphics[width=\linewidth]{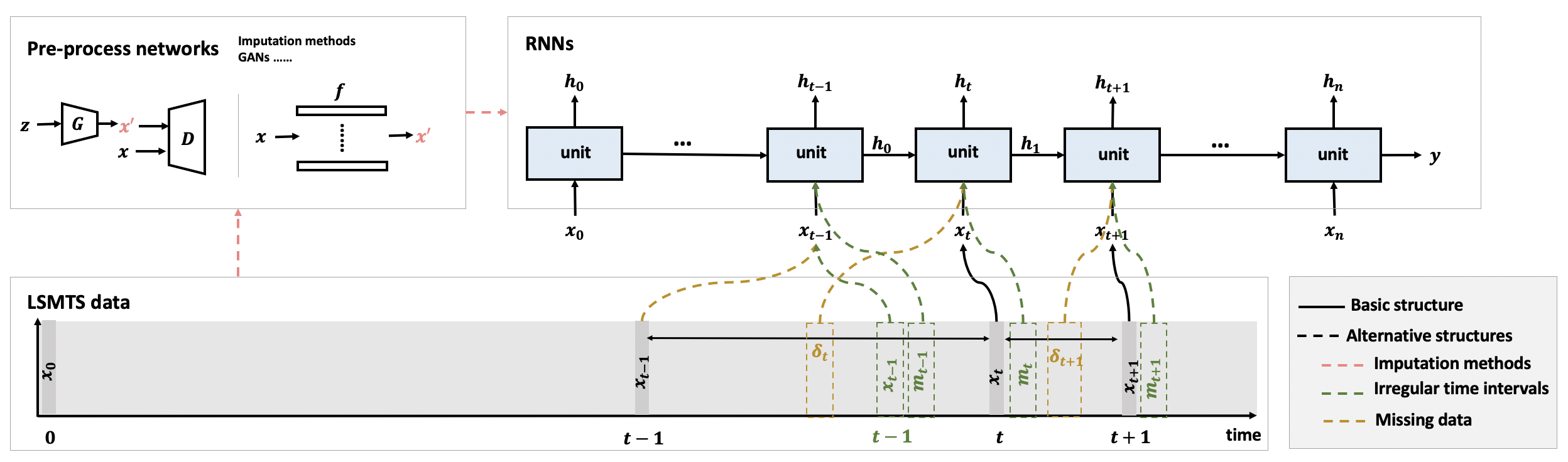}}
\caption{Structures of different RNN-based methods}
\label{fig:method_structures}
\end{figure*}

\subsubsection{Two-step approaches} \label{sec:twostep}

Two-step approaches ignore or impute missing values and then process downstream tasks based on the preprocessed data. A simple solution is to omit the missing data and perform analysis only on the observed data. But it can result in a large amount of useful data not being available \cite{1}. The core of these methods is how to impute the missing data.

Some basic methods are dedicated to filling the values, such as smoothing, interpolation \cite{19}, and spline \cite{20}. But they cannot capture the correlation between variables and complex patterns. Other methods estimate the missing values by spectral analysis \cite{21}, kernel methods \cite{22}, and expectation-maximization (EM) algorithm \cite{23}. However, simple reasoning design and necessary model assumptions make data imputation not accurate. Recently, with the vigorous development of deep learning, these methods have higher accuracy than traditional methods. RNNs and GANs mainly realize the deep learning-based data imputation methods. 

A substantial literature uses RNNs to impute the missing data in ISMTS. RNNs take sequence data as input, recursion occurs in the direction of sequence evolution, and all units are chained together. Their special structure endows them with processing sequence data by learning order dynamics. In a RNN, the current state $h_t$ is affected by the previous state $h_{t-1}$ and the current input $x_t$ and is described as 

\begin{equation}
h_t=\sigma(Wx_t+Uh_{t-1}+b)
\end{equation}

RNN can integrate basic methods, such as EM \cite{75} and linear model (LR) \cite{76}. The methods first estimate the missing values and again uses the re-constructed data streams as inputs to a standard RNN. However, EM imputes the missing values by using only the synchronous relationships across data streams (inter-series relations) but not the temporal relationships within streams (intra-series relations). LR interpolates the missing values by using only the temporal relationships within each stream (intra-series relations) but ignoring the relationships across streams (inter-series relations). Meanwhile, most of the RNN-based imputation methods, like simple recurrent network (SRN) and LSTM, which have been proved to be effective to impute medical data by Kim et al. \cite{62}, are also learn an incomplete relation with considering intra-series relations only.

Chu et al. \cite{26} have noticed the difference between these two relations in ISMTS data and designed multi-directional recurrent neural network (M-RNN) for both imputation and interpolation. M-RNN operates forward and backward in the intra-series directions according to an interpolation block  and operates across inter-series directions by an imputation block. They implanted imputation by a Bi-RNN structure recorded as function $\Phi$ and implanted interpolation by fully connected layers with function $\Psi$. The final objective function is mean squared error between the real data and calculated data.

\begin{equation}
\begin{aligned}
\Phi^{*},\Psi^{*} &=argmin_{\Phi,\Psi}\\
&=L(\{\Psi(\{x_t,\Phi(\{x_t,m,\delta\}_{t=1}^T)\})\}_{t=1}^T,X)
\end{aligned}
\end{equation}

Where $x,m$ and $\delta$ represent data value, masking and time interval we have defined in \ref{def:ISMTS}, we will not repeat it below. Bi-RNN is Bidirectional-RNN \cite{50}. It is an advanced RNN structure with forward and backward RNN chains. It have two hidden states for one time point in the above two orders. Two hidden states concatenate or sum into the final value in this time point. Unlike the basic Bi-RNN, the timing of inputs into the hidden layers of M-RNN is lagged in the forward direction and advanced in the backward direction. 

However, in M-RNN, the relations between missing variables are dropped, the estimated values are treated as constants which cannot be sufficiently updated. 

To solve the problem, Cao et al. \cite{27} proposed bidirectional recurrent imputation for time series (Brits) to predict missing values with bidirectional recurrent dynamics. In this model, the missing values are regarded as the variables in the model graph and get delayed gradients in both forward and backward directions with consistency constraints, which makes the estimation of missing values more accurate. It can update the predicted missing data with a combined three objective function $L$ – the errors of historical-based estimation $\hat{X}$, the feature-based estimation $\hat{Z}$ and the combined estimation $\hat{C}$, which not only considered the relations between missing data and known data, but also modeled the relations between missing data ignored by M-RNN.

\begin{equation}
L=L_e(X,\hat{X})+L_e(X,\hat{Z})+L_e(X,\hat{C})
\end{equation}

But Brits did not take both inter-series and intra-series relations into account, M-RNN solved it.

GANs are a type of deep learning model which train generative deep models through an adversarial process \cite{59}. From the perspective of game theory, GAN training can be seen as a minimax two-player game \cite{77} between generator $G$ and discriminator $D$ with the objective function. 

\begin{equation}
\begin{aligned}
&min_{G}max_{D}(V(D,G))\\
=&min_{G}max_{D}(E_{x\sim q(x)}(logD(x))+E_{z\sim q(z)}[log(1-D(G(z)))])   
\end{aligned}
\end{equation}

However, typical GANs require fully observed data during training. In response to this, Yoon et al. \cite{60} proposed generative adversarial imputation nets (GAIN) model. Different from the standard GAN, its generator receives both noise $Z$ and mask $M$ as input data, The masking mechanism makes missing data as input possible. GAIN's discriminator outputs both real and fake components. Meanwhile, A hint mechanism $H$ makes discriminator get some additional information in the form of a hint vector. GAIN changes the objective $min_{G}max_{D}(V(D,G))$ of basic GAN to

\begin{equation}
\begin{aligned}
min_{G}max_{D}(E_{Z,M,H}(MlogD(G(z,x,m),H)+(1-M)log(1-D(G(z,x,m),H))))
\end{aligned}
\end{equation}

To improve GAIN, Camino et al. \cite{78} used multiple-inputs and multiple-outputs to the generator and the discriminator. The method did the variable splitting by using dense layers connected parallelly for each variable. 

Zhang et al. \cite{79} designed Stackelberg GAN based on GAIN to impute the medical missing data for computational efficiency. Stackelberg GAN can generate more diverse imputed values by using multiple generators instead of a single generator and applying the ensemble of all pairs of standard GAN losses. 

\begin{equation}
min_{G}max_{D}(V(D,G))\to min_{G_{1},...,G_{I}}\sum_{k=1}^I max_{D}(V(D,G))
\end{equation}

The main goal of the above two-step methods is to estimate the missing values in the converted time series of ISMTS (convert irregularly sampled features to missing data features). However, in medical background, the ultimate goal is to carry out medical tasks such as mortality prediction \cite{31,52} and patient subtyping \cite{29,42,43}. Two separated steps may lead to the suboptimal analyses and predictions \cite{25} as the missing patterns are not effectively explored for final tasks. Thus, some researches proposed finding ways to solve the downstream tasks directly, rather than filling missing values.

\subsubsection{End-to-end approaches}\label{endtoend} 

End-to-end approaches process the downstream tasks directly based on modeling the time series with missing data. The core objective is to predict, classify, or clustering. Data imputation is an additional task or not even a task in this type of methods.

Lipton et al. \cite{38} demonstrated a simple strategy - using the basic RNN model to cope with missing data in sequential inputs and the output of RNN being the final characteristics for prediction. Then, to improve this basic idea, they addressed the task of multilabel classification of diagnoses by given clinical time series and found that RNNs can make remarkable use of binary indicators for missing data, improving AUC, and F1 significantly. Thus, they approached missing data by heuristic imputation directly model missingness as a first-class feature in the new work \cite{64}.

Similarly, Che at al. \cite{1} also use RNN idea to predict medical issues directly. For solving the missing data problem, they designed a kind of marking vector as the indicator for missing data. In this approach, the value $x$, the time interval $\delta$ and the masking $m$ impute missing data $x^*$ together. It first replaces missing data with the mean values, and then used the feedback loop to update the imputed values, which are the input of a standard RNN for prediction.

\begin{equation}
x^*\leftarrow[x,m,\delta]
\end{equation}

Meanwhile, they proposed GRU-Decay (GRU-D) to model EHRs data for medical predictions with trainable decays. The decay rate $\gamma$ weighs the correlation between missing data $x_t$ and other data (previous data $x_{t'}$ and mean data $\tilde{x}_t$). 

\begin{equation}
x_t\leftarrow m_tx_t+(1-m_t)\gamma x_{t'}+(1-m_t)(1-\gamma)\tilde{x}_t 
\end{equation}

Meanwhile, in this research, the authors plotted the Pearson correlation coefficient between variable missing rates of MIMIC-III dataset. They have observed that the missing rate is correlated with the labels, demonstrating the usefulness of missingness patterns in solving a prediction task. 

However, the above models \cite{1,6,27,63,64} are limited to using local information (empirical mean or the nearest observation) of ISMTS. For example, GRU-D assumed that a missing variable could be represented as the combination of its corresponding last observed value and the mean value. The global structure and statistics are not directly considered. The local statistics are unreliable when the continuous data misses (shown in Figure \ref{fig:important_missing}), or the missing rate rises up.

Tang et al. \cite{24} have realized this problem and designed LGnet, exploring the global and local dependencies simultaneously. They used GRU-D model local structure, grasping intra-series relations, and used a memory module to model the global structures, learning inter-series relations. The memory module $\mathcal{G}$ have $L$ rows, it capture the global temporal dynamics for missing values with the variable correlations $a$. Meanwhile, an adversarial training process can enhance the modeling of global temporal distribution.

\subsection{Raw data-based perspective} \label{sec:irregularity}

The alternative of processing the sequences with missing data by pre-discretizing ISMTS is constructing models which can directly receive ISMTS as input. The intuition of raw data-based perspective is from the characteristics of raw data itself - the intra -series relation and the inter-series relation. The intra -series relation of ISMTS is reflected in the irregular time intervals between two neighbor observations within one series; The inter-series relation is reflected in the different sampling rate of different time series. Thus, two subcategories are 1) irregular time intervals-based approaches and 2) multi-sampling rates-based approaches.

\subsubsection{Irregular time intervals-based approaches}\label{sec:irregulartime}

In EHRs setting, the time lapse between successive elements in patient records can vary from days to months, which is the characteristic of irregular time intervals in ISMTS. A better way to handle it is to model the unequally spaced data using time information directly.

Basic RNNs only process uniformly distributed longitudinal data by assuming that the sequences have an equal distribution of time differences. Thus, design of traditional RNNs may lead to suboptimal performance. 

For better RNN performance, Baytas et al. \cite{29} proposed a novel unit, time-aware LSTM (T-LSTM), to handle irregular time intervals in ISMTS for patient subtyping. T-LSTM is incorporated the elapsed time information into basic LSTM by a time decay function

\begin{equation}
g(\delta_{t})=\frac{1}{log(e+\delta_{t})}
\end{equation}

\noindent They applied a memory discount in coordination with elapsed time to capture the irregular temporal dynamics to adjust the hidden status $C_{t-1}$ of basic LSTM to a new hidden state $C_{t-1}^*$. 

\begin{equation}
\begin{aligned} 
C_{t-1}\to C_{t-1}^S\to \hat{C}_{t-1}^S,&C_{t-1}^T \to C_{t-1}^*;\\
C_{t-1}\to \tanh{(W_{d}C_{t-1}+b_{d})}\to C_{t-1}^S\cdot g(\delta_{t}),&C_{t-1}-C_{t-1}^S\to C_{t-1}^T-\hat{C}_{t-1}^S.
\end{aligned}
\end{equation}

However, when ISMTS is univariate, T-LSTM is not a completely irregular time intervals-based method. For the multivariate ISMTS, it has to align multiple time series and filling missing data first. Where they have to solve the missing data problem again. But the research did not mention the specific filling strategy and used simple interpolation like mean values when data preprocessing.

For the multivariate ISMTS and the alignment problem, Tan et al. \cite{54} gave an end-to-end dual-attention time-aware gated recurrent unit (DATA-GRU) to predict patients' mortality risk. DATA-GRU uses a time-aware GRU structure T-GRU as same as T-LSTM. Besides, the authors give the strategy of multivariate data alignment problem. When aligning different time series to multi dimensions, previous missing data approaches, such as GRU-D \cite{1} and LGnet \cite{24}, assigned equal weights to observed data and imputed data, ignoring the relatively larger unreliability of imputation compared with actuality. DATA-GRU tackles this difficulty by a novel dual-attention structure - unreliability-aware attention $\alpha^u$ with reliability score $c$ and symptom-aware attention $\alpha^s$. The dual-attention structure jointly considers the data-quality and the medical-knowledge. 

\begin{equation}
\begin{aligned} 
\alpha^u=sigmoid(Wc_t+b),\quad &\alpha^s=TGRU(W\alpha_{0/1}^s)\\
x^u=x\odot \alpha^u,\quad &x^s=x\odot \alpha^s
\end{aligned}
\end{equation}

Further, the attention-like structure makes DATA-GRU explainable according to the interpretable embedding, which is an urgently needed issue in medical tasks.

Instead of using RNNs to learn the order dynamics in ISMTS, Bahadori et al. \cite{63} have proposed methods for analyzing multivariate clinical time series that are invariant to temporal clustering. The events in EHRs may appear in a single admission together or may disperse over multiple admissions. For example, the authors postulated that whether a series of blood tests are completed at once or in rapid succession should not alter predictions. Thus, they designed a data augmentation technique, temporally coarsening, to exploits temporal-clustering invariance to regularize deep neural networks optimized for clinical prediction tasks. Moreover, they proposed a multi-resolution ensemble (MRE) model with the coarsening transformed inputs to improve predictive accuracy.

\subsubsection{Multi sampling rates-based approaches}\label{sec:multisampling}

Only modeling the irregular time intervals of intra-series relation would ignore the multi-sampling rate relation of inter-series relation. Further, modeling inter-series relation is also a reflection of considering the global structure of ISMTS. 

The above RNN-based methods of irregular time intervals-based category only consider the local order dynamics information. Although LGnet \cite{24} has integrated the global structures, it incorporates all of the information from all time points into an interpolation model, which is redundant and low adaptive. Some models can also learn the global structures of time series, like a basic model Kalman filters \cite{83} and a deep learning deep Markov models \cite{84}. However, this kind of models mainly process the every time series with a stable sampling rate.

Che et al. \cite{56} focused on the problem of modeling multi-rate multivariate time series and proposed a multi-rate hierarchical deep Markov model (MR-HDMM) for healthcare forecasting and interpolation tasks. MR-HDMM learns generation model and inference network by auxiliary connections and learnable switches. The latent hierarchical structure reflected in the states/switches $s$ factorizing by joint probability $p$ with layer $z$.

\begin{equation}
p(x_1,z_1,s_1 |z_0 )=p(x_1 |z_1 )p(z_1,s_1 |z_0 )
\end{equation}

\begin{equation}
a=\sum_{l=1}^{L}f(\mathcal{G})^l\mathcal(G)(l)
\end{equation}

\begin{figure*}[!t]
\centering{
\includegraphics[width=0.6\linewidth]{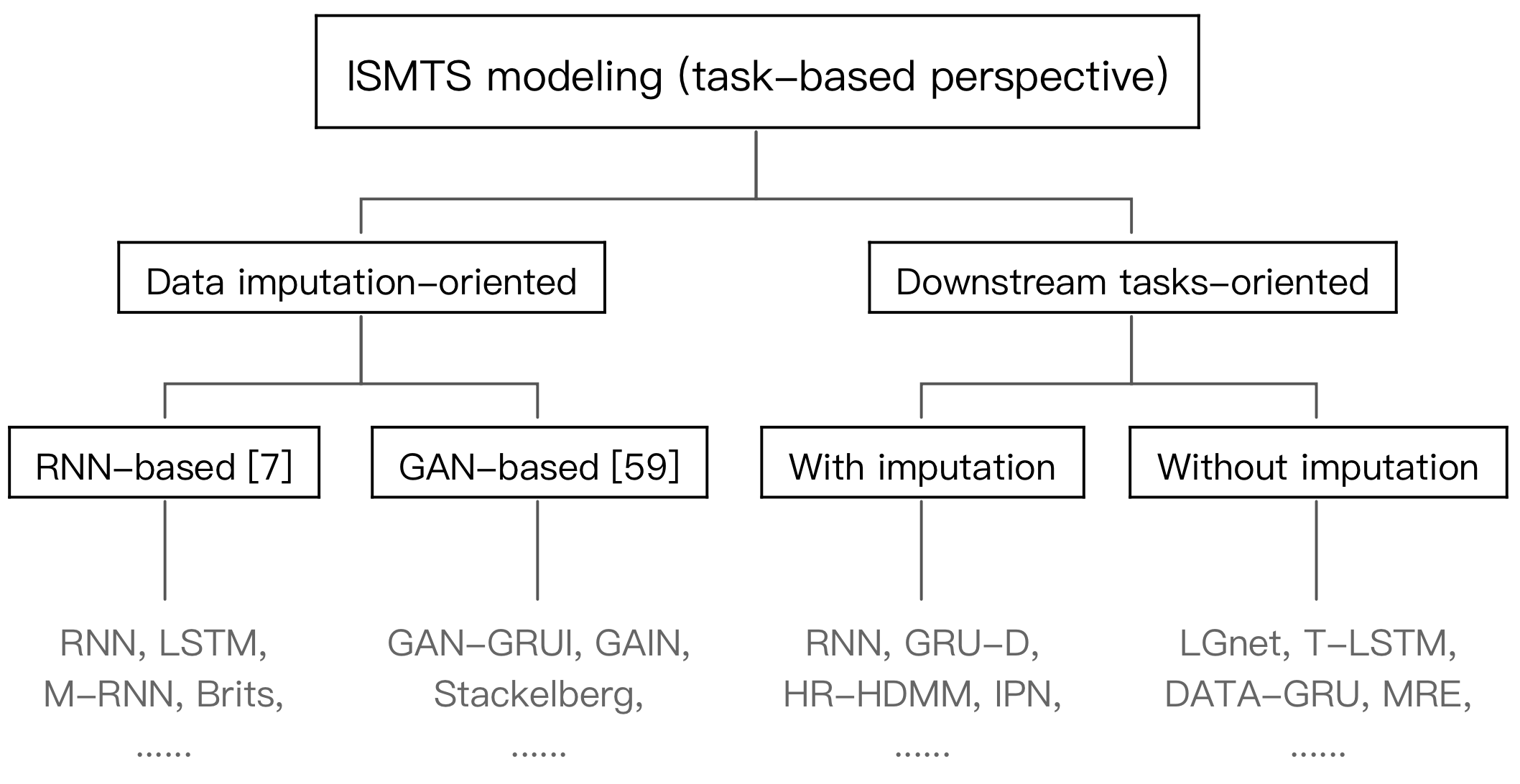}}
\caption{Categories based on task-driven}
\label{fig:task_driven}
\end{figure*}

These structures can capture the temporal dependencies and data generation process. Similarly, Binkowski et al. \cite{81} presented an autoregressive framework for regression tasks by modeling ISMTS data. The core idea of implementation is roughly similar with MR-HDMM.

However, these methods considered the different sampling rates between series but ignored the irregular time intervals in each series. They process the data with a stable sampling rate (uniform time intervals) for each time series. For the stable sampling rate, they have to use forward or linear interpolation, where the global structures are omitted again for getting the uniform intervals. The Gaussian process can build global interpolation layers for process multi-sampling rate data. Li et al. \cite{80} and Futoma et al. \cite{82} used this technique. But if a time series is multivariate data, covariance functions are challenging due to the complicated and expensive computation.

Satya et al. \cite{55} designed a fully modular interpolation-prediction network (IPN). IPN has an interpolation network to accommodate the complexity of ISMTS data and provide the multi-channel output by modeling three information - broad trends $\mathcal{\chi}$, transients $\tau$ and local observation frequencies $\lambda$. The three information is calculated by a low-pass interpolation $\theta$, a high-pass interpolation $\gamma$ and an intensity function $\lambda$.

\begin{equation}
\begin{aligned}
\chi=h^\chi(\sigma,\lambda)&,\quad \tau=h^\tau(\gamma,\chi);\\
\sigma=h^\sigma(\chi,\tau,r),\quad \gamma=&h^\gamma(\chi,\tau,r),\quad \lambda=h^\lambda(\chi,\tau,r)
\end{aligned}
\end{equation}

IPN also has a prediction network which operates the regularly partitioned inputs from the former interpolation module. In addition to taking care of data relationships from multiple perspectives, IPN can make up for the lack of modularity in \cite{1} and address the difficulty of the complexity of the Gaussian process interpolation layers in \cite{80,82}.

\section{Categorization based on task-driven} \label{sec:categories2}

Modeling ISTS data aims to achieve two main tasks: 1) Missing data imputation and 2) Downstream tasks. The specific categories are shown in Figure \ref{fig:task_driven}.

Missing data imputation is of practical significance, as works on machine learning have become actively, getting large amounts of complete data has become an important issue. However, it is almost impossible in the real world to get complete data for many reasons like lost records. In many cases, the time series with missing values becomes useless and then thrown away. This results in a large amount of data loss. The incomplete data has adverse effects when learning a model \cite{59}. 

Existing basic methods, such as interpolation \cite{19} kernel methods \cite{22} and EM algorithm \cite{23,75}, have been proposed a long time ago. With the popularity of deep learning in recent years, most new methods are implemented by artificial neural networks (ANNS). One of the most popular models is RNN \cite{7}. RNNs can capture long-term temporal dependencies and use them to estimate the missing values. Existing works \cite{16,27,62,64,65,68} have designed several special RNN structures to adapt the missingness and achieve good results. Another popular model is GANs \cite{66}, which generate plausible fake data through adversarial training. GAN has been successfully applied to face completion and sentence generation \cite{67,68,69,70}. Based on their data generation abilities, some research \cite{13,59,60,61} have applied GAN on time series data generation with considering sequence information into the process.

Downstream tasks generally include prediction, classification, and clustering. For ISMTS data, medical prediction (such as mortality prediction, disease classification and image classification) \cite{31,37,38,39,52}, concept representation \cite{41,42} and patient typing \cite{29,42,43} are three main tasks. The downstream task-oriented methods calculate missing values and perform downstream tasks simultaneously, which is expected to avoid suboptimal analyses and predictions caused by the not effectively explored missing patterns due to the separation of imputations and final tasks \cite{25}. Most methods \cite{1,24,29,54,55,56,63,64} use deep learning technology to achieve higher accuracy on tasks.

\section{Experiments} \label{sec:experiments}

In this section, we apply the above methods on four datasets and two tasks. We will analyze the method through the experimental results.

\subsection{Datasets} \label{sec:datasets}

Four datasets were used to evaluate the performance of baselines.

\textbf{MIMIC-III dataset} \cite{32} is a freely accessible de-identified critical care database, developed and maintained by the Massachusetts Institute of Technology Laboratory for Computational Physiology. In this work, we extract diagnoses and demographics information for disease prediction and choose records with more than one visit. The dataset comprises 19,993 hospital admissions, 260,326 diagnoses and 4,893 unique ICD-9 codes of 7,537 patients. The average number of visits per patient is 2.66. There are 13.02 codes per visit on average and up to 39 codes in a visit.

\textbf{CINC2012 dataset} \cite{31} consist of records from 12,000 ICU stays and have 4000 multivariate clinical time series. All patients were adults who were admitted for a wide variety of reasons to cardiac, medical, surgical, and trauma ICUs. Each record is a multivariate time series of roughly 48 hours and contains 41 variables such as Albumin, heart rate, glucose etc.

\textbf{CINC2019 dataset} \cite{91} is publicly available and comes from two hospitals; it contains 30,336 patient admission records and 2,359 records of diagnosed sepsis cases. It is a set of multivariate time series that contains 40 related features, 8 kinds of vital signs, 26 kinds of laboratory values and 6 kinds of demographics. The time interval is 1 hour. The sequence length is between 8 and 336, and 29,414 records have lengths less than 60.

\textbf{COVID-19 dataset} \cite{57} is collected between 10 January and 18 February 2020 from Tongji Hospital of Tongji Medical College, Huazhong University of Science and Technology, Wuhan, China. The dataset contains 375 patients with 6120 blood sample records as training set, 110 patients with 757 records as test set and 80 characteristics.

\subsection{Experiment setting} \label{sec:Experimentsetting}

The experiments have two tasks – 1) mortality prediction and 2) data imputation. The mortality prediction tasks use the time series of 48 hours before onset time from the above four datasets. The imputation tasks use 8 features (using the method in \cite{56}) which are eliminated 10\% of observed measurements from data. The eliminated data is the new ground-truth.

For RNN-based method, we fix the dimension of hidden state is 64. For GAN-based methods, the series inputs also use RNN structure. For final prediction, all methods use one 128-dimensions FC layer and one 64-dimensions FC layer. All methods apply Adam Optimizer \cite{85} with $\alpha=0.001$, $\beta_1=0.9$ and $\beta_2=0.999$. We use the learning rate decay $\alpha_{current}= \alpha_{initial}\cdot \gamma^{\frac{\textit{global step}}{\textit{decay steps}}}$ with decay rate  $\gamma=0.98$ and the decay step is 2000. The 5-fold cross validation is used for both two tasks.

\subsection{Evaluation metrics}

The prediction results were evaluated by assessing the area under curve of receiver operating characteristic (AUC-ROC). ROC is a curve of true positive rate (TPR) and false positive rate (FPR). TN, TP, FP and FN stand for true positive, true negative, false positive and false negative rates.

\begin{equation}
\begin{aligned}
TPR=\frac{TP}{TP+FN}\\
FPR=\frac{FP}{TN+FP}
\end{aligned}
\end{equation}

We evaluate the imputation performance in terms of mean squared error (MSE). For $i$th item, $\hat{x}_i$ is the real value and $x_i$ is the predicting value. The number of missing values is $N$.
\begin{equation}
MSE=\frac{\sum_{i=1}^{N}(x_i-\hat{x_i})^2}{n}
\end{equation}

\begin{table*}[t]
\caption{Performance comparison for mortality prediction task (in AUC-ROC)}\label{tb:exp1}
\centering
\begin{tabular}{lllll}
\hline
 &MIMIC-III	&CINC2012	&CINC2019	&COVID-19\\
\hline
RNN\cite{64}	&0.809 ± 0.014	&0.800 ± 0.016	&0.825 ± 0.024	&0.945 ± 0.004\\
\hline
LSTM\cite{62}	&0.812 ± 0.009	&0.805 ± 0.010	&0.829 ± 0.019	&0.945 ± 0.005\\
\hline
GRU-D\cite{1}	&0.829 ± 0.003	&0.818 ± 0.009	&0.835 ± 0.013	&\textbf{0.965 ± 0.003}\\
\hline
M-RNN\cite{26}	&0.827 ± 0.005	&0.820 ± 0.011	&0.842 ± 0.010	&0.959 ± 0.003\\
\hline
Brits\cite{27}	&\textbf{0.833 ± 0.002}	&0.819 ± 0.012	&0.839 ± 0.013	&0.959 ± 0.002\\
\hline
T-LSTM\cite{29}	&0.817 ± 0.004	&0.804 ± 0.010	&0.831 ± 0.014	&\underline{0.963 ± 0.003}\\
\hline
DATA-GRU\cite{54}	&0.832 ± 0.006	&\underline{0.822 ± 0.012}	&\textbf{0.851 ± 0.012}	&0.961 ± 0.003\\
\hline
LGnet\cite{24}	&\underline{0.833 ± 0.003}	&\underline{0.822 ± 0.013}	&0.843 ± 0.013	&0.956 ± 0.002\\
\hline
IPN \cite{55}	&0.831 ± 0.003	&\textbf{0.824 ± 0.009}	&\underline{0.844 ± 0.015}	&0.960 ± 0.003\\
\hline
\end{tabular}
\end{table*}

\begin{table*}[t]
\caption{Performance comparison for imputation task (in RMSE)}\label{tb:exp2}
\centering
\begin{tabular}{lllll}
\hline
 &MIMIC-III	&CINC2012	&CINC2019	&COVID-19\\
\hline
RNN\cite{64}	&4.985	&4.180	&2.901	&1.685\\
\hline
LSTM\cite{62}	&4.712	&4.046	&2.899	&1.710\\
\hline
GRU-D\cite{1}	&4.412	&3.567	&2.379	&1.543\\
\hline
M-RNN\cite{26}	&4.435	&3.236	&2.337	&\textbf{1.530}\\
\hline
Brits\cite{27}	&4.339	&3.238	&2.439	&\underline{1.533}\\
\hline
GAN-GRUI\cite{13}	&\textbf{3.920}	&\underline{3.129}	&\underline{2.225}	&\underline{1.533}\\
\hline
Stackelberg\cite{79} &4.431	&3.741	&2.435	&1.932\\
\hline
MR-HDMM\cite{56}	&\underline{4.015}	&\textbf{3.113}	&\textbf{2.217}	&1.555\\
\hline
\end{tabular}
\end{table*}

\begin{table*}[t]
\caption{Performance comparison for mortality prediction task based on the imputation data (in AUC-ROC)}\label{tb:exp3}
\centering
\begin{tabular}{lllll}
\hline
 &MIMIC-III	&CINC2012	&CINC2019	&COVID-19\\
\hline
RNN\cite{64}	&0.805	&0.783	&0.814	&0.945\\
\hline
LSTM\cite{62}	&0.802	&0.794	&0.819	&0.943\\
\hline
GRU-D\cite{1}	&0.819	&0.813	&0.840	&\textbf{0.949}\\
\hline
M-RNN\cite{26}	&\underline{0.826}	&\textbf{0.818}	&\textbf{0.842}	&0.946\\
\hline
Brits\cite{27}	&0.823	&0.810	&0.831	&\underline{0.947}\\
\hline
GAN-GRUI\cite{13}	&0.820	&0.809	&\textbf{0.842}	&0.945\\
\hline
Stackelberg\cite{79} &0.798	&0.767	&0.789	&0.941\\
\hline
MR-HDMM\cite{56}	&\textbf{0.829}	&\underline{0.814}	&\underline{0.841}	&0.945\\
\hline
\end{tabular}
\end{table*}

\subsection{Results} \label{sec:results}

Table 1 shows the performances of baselines for the mortality prediction task. For the two categories of technology-driven methods, each has its own merits, but irregularity-based methods work relatively well. Missing data-based methods have 2/4 top 1 results and 2/5 top 2 results, while irregularity-based methods have 2/4 top 1 results and 3/5 top 2 results. 

For the methods of whether the two series relation are considered, the methods that take both inter-series relation and intra-series relation (both global and local structures) into account perform better. IPN, LGnet, and DATA-GRU have relatively good results. 
For different datasets, the methods show different effects. For example, as COVID-19 is a small dataset, unlike the other three datasets, the relatively simple methods perform better on this dataset, like T-LSTM, which doesn't perform very well on the other three datasets.

Table 2 shows the performances of baselines for the imputation prediction task. The performances of RNN-based methods and GAN-based methods are similar. RNN-based methods have 3/4 top 1 results and 2/5 top 2 results, GAN-based methods have 1/4 top 1 results and 3/5 top 2 results. MR-HDMM and GAN-GRUI are the two best methods. 

The data imputation is better in the Sepsis and COVID-19 dataset. Perhaps the time series in these two datasets is from the patients who suffered from the same disease. That's probably why they also have relatively better results in the prediction task.

Table 3 shows a basic RNN model's performance for mortality prediction tasks based on baselines' imputation data. Different from the results in Table 2, the RNN-based methods perform better. Where the RNN-based methods have 4/5 top 1 results, but GAN-based methods have 1/5. The reason may be that the RNN-based approaches have integrated the downstream tasks when imputing. So, the data generated by them is more suitable for the final prediction task.

\section{Discussions} \label{sec:discussions}

According to the analysis of technologies and experiment results, in this section, we will discuss ISMTS modeling task from three perspectives - 1) imputation task with prediction task, 2) intra-series relation with inter-series relation / local structure with global structure and 3) missing data with raw data. The conclusions of the approaches in this survey are in Table \ref{tb:conclusions}.

\subsection{Challenges} \label{sec:challenges}

Based on the above five perspectives, we summarize the challenges as follows.

How to balance the imputation with the prediction? Different kinds of methods suit different tasks. GANs prefer imputation while RNNs prefer prediction. However, in the medical setting, aiming at different datasets, the conclusion does not seem correct. For example, missing data is generated better by RNNs than GANs in the COVID-19 dataset. And the two-step methods based on GANs for mortality prediction are no worse than using RNNs directly. Therefore, it seems difficult to achieve a general and effective modeling method in medical settings. The method should be specified according to the specific task and the characteristics of the datasets.

How to handle the intra-series relation with inter-series relation of ISMTS? In other words, how to trade off the local structure with global structure. In ISMTS format, a patient has several time series of vital signs connected to the diagnoses of diseases or the probability of death. Seeing these time series as a whole multivariate data sample, intra-series relations are reflected in longitudinal dependences and horizontal dependencies. The longitudinal dependencies contain the sequence order and context, time intervals, and decay dynamics. The horizontal dependence is the relations between different dimensions. And the inter-series relations are reflected in the patterns of time series of different samples. 

However, when seeing these time series as separated multi-samples of a patient, the relations will change. Intra-series relations change to the dependencies of values observed on different time steps in a univariate ISMTS. The features of different time intervals should be taken care of. Inter-series relations change to the pattern relations between different patients' different samples and between different time series of the same vital sign. 

For the structural level, modeling intra-series relations is basically at the local level, while modeling inter-series relations is global. It is not clear what kind of consideration and which structure will make the results better. Modeling local and global structures seems to perform better in morality prediction, but it is a more complex method, and it's not universal for different datasets.

How to choose the modeling perspective, missing data-based or irregularity-based? Both two kinds of methods have advantages and disadvantages. Most existing works are missing data-based and there are methods of estimating missing data for a long time \cite{86}. In settings of missing data-based perspective, the discretization interval length is a hyper-parameter needs to be determined. If the interval size is large, missing data is less, but several values will show in the same interval; If the interval size is small, the missing data becomes more. No values in an interval will hamper the performance, while too many values in an interval need an ad-hoc choosing method. Meanwhile, missing data-based methods have to interpolate new values, which may artificially introduce some naturally occurring dependencies. Over-imputation may result in an explosion in size and the pursuit of multivariate data alignment may lead to the loss of raw data dependency. Thus, of particular interest are irregularity-based methods that can learn directly by using multivariate sparse and irregularly sampled time series as input without the need for other imputation.

However, although the raw data-based methods have metrics of no artificial dependencies introduced, they suffer from not achieving the desired results, complex designs, and large parameters. Irregular time intervals-based methods are not complex as they can be achieved by just injecting time decay information. But in terms of specific tasks, such as morality prediction, the methods seem not as good as we think (concluded from experiments section). Meanwhile, for multivariable time series, these methods have to align values on different dimensions, which leads to missing data problems again. Multi-sampling rates-based methods will not cause missing data. However, processing multiple univariate time series at the same time requires more parameters and is not friendly to batch learning. Meanwhile, modeling the entire univariate series may require data generation model assumptions. 

\subsection{Opportunities}

Considering the complex patient states, the amount of interventions and the real-time requirement, the data-driven approaches by learning from EHRs are the desiderata to help clinicians.

Although some difficulties have not been solved yet, the deep learning method does show a better ability to model medical ISMTS data than the basic methods. Basic methods can't model ISMTS completely as interpolation-based methods \cite{19,20} just exploit the correlation within each series, imputation-based methods \cite{23,87} just exploit the correlation among different series, matrix completion-based methods \cite{88,89} assume that the data is static and ignore the temporal component of the data. Deep learning methods use parameter training to learn data structures, and many basic methods can be integrated into the designs of neural networks. The deep learning methods introduced in this survey basically solve the problem of common methods and have achieved state-of-the-art in medical prediction tasks, including mortality prediction, disease prediction, and admission stay prediction. Therefore, the deep learning model based on ISMTS data has a broad prospect in medical tasks.

The deep learning methods, both RNN-based and GAN-based methods mentioned in this survey, are troubled by poor interpretability \cite{90,111}, and clinical settings prefer interpretable models. Although this defect is difficult to solve due to models' characteristics, some researchers have made some breakthroughs and progress. For example, the attention-like structures which are used in \cite{37,39} can give an explanation for medical predictions.

\begin{table*}[t]
\caption{Conclusions of introduced methods for medical tasks using EHRs data}\label{tb:conclusions}
\centering
\begin{tabular}{|l|l|l|l|l|}
\hline
Category &Subcategory &Researches &Advantages &Disadvantages\\
\hline
\multirow{2}{*}{Missing data-based}  & \shortstack[l]{Two-step} &\shortstack[l]{\cite{7,13,16,27}\\\cite{59,60,61,62}\\\cite{79,120,122}} &\shortstack[l]{Generality;\\ Ability of imputation.} &\shortstack[l]{Suboptimal prediction; \\Incomplete data relation; \\Data generation patterns \\assumptions; \\Introduced artificial \\dependency.}\\
\cline{2-5}
&End-to-end &\shortstack[l]{\cite{1,24,64,113}} &\shortstack[l]{Optimal prediction} &\shortstack[l]{ Non-commonality;\\ Introduced artificial \\dependency.}\\
\hline
\multirow{2}{*}{Raw data-based} &\shortstack[l]{Irregular time\\ intervals-based} &\shortstack[l]{\cite{29,54,63,112}\\\\ \cite{114,117,118}\\ \cite{119,123}} & \shortstack[l]{No artificial dependency.} & \shortstack[l]{Low-applicability for \\multivariate data;\\ Incomplete data \\relation.}\\
\cline{2-5}
&\shortstack[l]{Multi sampling\\ rates-based} &\shortstack[l]{\cite{55,56,115,116}} &\shortstack[l]{No artificial dependency;\\No data imputation.} & \shortstack[l]{Implementation complexity;\\Data generation \\patterns assumptions.}\\
\hline
\end{tabular}
\end{table*}

\section{Conclusion} \label{sec:conclusion}

This survey introduced a kind of data – irregularly sampled medical time series (ISMTS). Combined with medical settings, we described characteristics of ISMTS. Then, we have investigated the relevant methods for modeling ISMTS data and classified them by technology-driven perspective and task-driven perspective. For each category, we divided the subcategories in detail and represented each specific model's implementation method. Meanwhile, according to imputation and prediction experiments, we analyzed the advantages and disadvantages of some methods and made conclusions. Finally, we summarized the challenges and opportunities of modeling ISMTS data task.

\bibliographystyle{unsrt} 
\bibliography{ISTS}

\begin{thebibliography}{100}

\bibitem{1}
Zhengping Che, Sanjay Purushotham, Kyunghyun Cho, David Sontag, and Yan Liu.
\newblock Recurrent neural networks for multivariate time series with missing
  values.
\newblock {\em Sentific Reports}, 8(1):6085, 2018.

\bibitem{2}
Xingjian Shi, Zhourong Chen, Hao Wang, Dit{-}Yan Yeung, Wai{-}Kin Wong, and
  Wang{-}chun Woo.
\newblock Convolutional {LSTM} network: {A} machine learning approach for
  precipitation nowcasting.
\newblock In {\em Advances in Neural Information Processing Systems}, pages
  802--810, 2015.

\bibitem{3}
Xian Wu.
\newblock Restful: Resolution-aware forecasting of behavioral time series data.
\newblock In {\em the 27th ACM International Conference}, 2018.

\bibitem{8}
Dongkuan Xu, Wei Cheng, Bo~Zong, Dongjin Song, and Xiang Zhang.
\newblock Tensorized lstm with adaptive shared memory for learning trends in
  multivariate time series.
\newblock {\em Proceedings of the AAAI Conference on Artificial Intelligence},
  34(2):1395--1402, 2020.

\bibitem{9}
M.~Ali, A.~Alqahtani, M.~W. Jones, and X.~Xie.
\newblock Clustering and classification for time series data in visual
  analytics: A survey.
\newblock {\em IEEE Access}, PP(99):1--1, 2019.

\bibitem{10}
Ziqiang Cheng, Yang Yang, Wei Wang, Wenjie Hu, Yueting Zhuang, and Guojie Song.
\newblock Time2graph: Revisiting time series modeling with dynamic shapelets.
\newblock In {\em The Thirty-Fourth {AAAI} Conference on Artificial
  Intelligence}, pages 3617--3624, 2020.

\bibitem{11}
Karan Aggarwal, Shafiq~R. Joty, Luis Fern{\'{a}}ndez{-}Luque, and Jaideep
  Srivastava.
\newblock Adversarial unsupervised representation learning for activity
  time-series.
\newblock In {\em The Thirty-Third {AAAI} Conference on Artificial
  Intelligence}, pages 834--841, 2019.

\bibitem{4}
Huaxiu Yao, Xianfeng Tang, Hua Wei, Guanjie Zheng, and Zhenhui Li.
\newblock Revisiting spatial-temporal similarity: A deep learning framework for
  traffic prediction.
\newblock {\em Proceedings of the AAAI Conference on Artificial Intelligence},
  33:5668--5675, 2019.

\bibitem{36}
Benjamin Shickel, Patrick Tighe, Azra Bihorac, and Parisa Rashidi.
\newblock Deep {EHR:} {A} survey of recent advances in deep learning techniques
  for electronic health record {(EHR)} analysis.
\newblock {\em {IEEE} J. Biomed. Health Informatics}, 22(5):1589--1604, 2018.

\bibitem{31}
Ivanovitch Silva, Galan Moody, Daniel~J Scott, Leo~A Celi, and Roger~G Mark.
\newblock Predicting in-hospital mortality of icu patients: The
  physionet/computing in cardiology challenge 2012.
\newblock {\em Critical Care Medicine}, 2012.

\bibitem{52}
Shenda Hong, Yanbo Xu, Alind Khare, Satria Priambada, Kevin~O. Maher, Alaa
  Aljiffry, Jimeng Sun, and Alexey Tumanov.
\newblock {HOLMES:} health online model ensemble serving for deep learning
  models in intensive care units.
\newblock In {\em The 26th {ACM} {SIGKDD} Conference on Knowledge Discovery and
  Data Mining}, pages 1614--1624, 2020.

\bibitem{37}
Fenglong Ma, Radha Chitta, Jing Zhou, Quanzeng You, Tong Sun, and Jing Gao.
\newblock Dipole: Diagnosis prediction in healthcare via attention-based
  bidirectional recurrent neural networks.
\newblock In {\em Proceedings of the 23rd {ACM} {SIGKDD} International
  Conference on Knowledge Discovery and Data Mining}, pages 1903--1911, 2017.

\bibitem{38}
Zachary~Chase Lipton, David~C. Kale, Charles Elkan, and Randall~C. Wetzel.
\newblock Learning to diagnose with {LSTM} recurrent neural networks.
\newblock In {\em 4th International Conference on Learning Representations,
  {ICLR}}, 2016.

\bibitem{39}
Edward Choi, Mohammad~Taha Bahadori, Jimeng Sun, Joshua Kulas, Andy Schuetz,
  and Walter~F. Stewart.
\newblock {RETAIN:} an interpretable predictive model for healthcare using
  reverse time attention mechanism.
\newblock In {\em Advances in Neural Information Processing Systems}, pages
  3504--3512, 2016.

\bibitem{41}
Edward Choi, Mohammad~Taha Bahadori, Elizabeth Searles, Catherine Coffey,
  Michael Thompson, James Bost, Javier Tejedor{-}Sojo, and Jimeng Sun.
\newblock Multi-layer representation learning for medical concepts.
\newblock In {\em Proceedings of the 22nd {ACM} {SIGKDD} International
  Conference on Knowledge Discovery and Data Mining}, pages 1495--1504, 2016.

\bibitem{42}
Edward Choi, Cao Xiao, Walter~F. Stewart, and Jimeng Sun.
\newblock Mime: Multilevel medical embedding of electronic health records for
  predictive healthcare.
\newblock In {\em Advances in Neural Information Processing Systems 31: Annual
  Conference on Neural Information Processing Systems}, pages 4552--4562, 2018.

\bibitem{29}
Inci~M. Baytas, Cao Xiao, Xi~Zhang, Fei Wang, and Jiayu Zhou.
\newblock Patient subtyping via time-aware lstm networks.
\newblock In {\em Acm Sigkdd International Conference on Knowledge Discovery
  and Data Mining}, 2017.

\bibitem{43}
Zhengping Che, David~C. Kale, Wenzhe Li, Mohammad~Taha Bahadori, and Yan Liu.
\newblock Deep computational phenotyping.
\newblock In {\em Proceedings of the 21th {ACM} {SIGKDD} International
  Conference on Knowledge Discovery and Data Mining}, pages 507--516, 2015.

\bibitem{93}
W.~Young, G.~Weckman, and W.~Holland.
\newblock A survey of methodologies for the treatment of missing values within
  datasets: limitations and benefits.
\newblock {\em Theoretical Issues in Ergonomics Science}, 12(1):p.15--43, 2011.

\bibitem{94}
G.~H. Golub.
\newblock Singular value decomposition and least squares solutions.
\newblock {\em Numerische Mathematik}, 14(5):403--420, 1970.

\bibitem{95}
Wang~Wei Dong and Zhenjiang.
\newblock An efficient nearest neighbor classifier algorithm based on
  pre-classify.
\newblock {\em Computer ence}, 2007.

\bibitem{96}
Katherine Godfrey.
\newblock Simple linear regression in medical research.
\newblock {\em N Engl J Med}, 315(26):1629--1636, 1986.

\bibitem{97}
Mohammed Khalilia, Sounak Chakraborty, and Mihail Popescu.
\newblock Predicting disease risks from highly imbalanced data using random
  forest.
\newblock {\em Bmc Medical Informatics and Decision Making}, 11(1):51--51,
  2011.

\bibitem{98}
Zhang and Guojun.
\newblock A modified svm classifier based on rs in medical disease prediction.
\newblock {\em International Journal of Advanced Research in Computer
  Engineering and Technology}, pages 144--147, 2009.

\bibitem{100}
Ruoxuan Cui, Manhua Liu, and Alzheimer's Disease~Neuroimaging Initiative.
\newblock Rnn-based longitudinal analysis for diagnosis of alzheimer's disease.
\newblock {\em Comput. Medical Imaging Graph.}, 73:1--10, 2019.

\bibitem{101}
Wang Yueming, Lin Kang, Qi~Yu, Lian Qi, Feng Shaozhe, Pan Gang, and Wu~Zhaohui.
\newblock Estimating brain connectivity with varying-length time lags using a
  recurrent neural network.
\newblock {\em IEEE Transactions on Biomedical Engineering}, 65:1953--1963,
  2018.

\bibitem{102}
Huilong Duan, Zhoujian Sun, Wei Dong, Kunlun He, and Zhengxing Huang.
\newblock On clinical event prediction in patient treatment trajectory using
  longitudinal electronic health records.
\newblock {\em IEEE Journal of Biomedical and Health Informatics},
  24(7):2053--2063, 2020.

\bibitem{103}
Joonnyong Lee, Sukkyu Sun, Seung~Man Yang, Jang~Jay Sohn, Jonghyun Park, Saram
  Lee, and Hee~Chan Kim.
\newblock Bidirectional recurrent auto-encoder for photoplethysmogram
  denoising.
\newblock {\em IEEE Journal of Biomedical and Health Informatics},
  23(6):2375--2385, 2019.

\bibitem{104}
Yang~Zhen Yu and Jing Hui.
\newblock A deep learning method based on hybrid auto-encoder model.
\newblock {\em IEEE Journal of Biomedical and Health Informatics}, pages
  1100--1104, 2017.

\bibitem{105}
Mengting Chai and Yuanping Zhu.
\newblock Research and application progress of generative adversarial networks.
\newblock {\em Computer Engineering}, 2019.

\bibitem{106}
Bing Yan, Haoqian Wang, Wang Xingzheng, and Zhang Yongbing.
\newblock An accurate saliency prediction method based on generative
  adversarial networks.
\newblock In {\em 2017 IEEE International Conference on Image Processing
  (ICIP)}, 2017.

\bibitem{24}
Xianfeng Tang, Huaxiu Yao, Yiwei Sun, Charu~C. Aggarwal, Prasenjit Mitra, and
  Suhang Wang.
\newblock Joint modeling of local and global temporal dynamics for multivariate
  time series forecasting with missing values.
\newblock In {\em The Thirty-Fourth {AAAI} Conference on Artificial
  Intelligence}, pages 5956--5963, 2020.

\bibitem{64}
Zachary~C. Lipton, David~C. Kale, and Randall~C. Wetzel.
\newblock Directly modeling missing data in sequences with rnns: Improved
  classification of clinical time series.
\newblock In {\em Proceedings of the 1st Machine Learning in Health Care},
  volume~56, pages 253--270, 2016.

\bibitem{27}
Wei Cao, Dong Wang, Jian Li, Hao Zhou, Lei Li, and Yitan Li.
\newblock {BRITS:} bidirectional recurrent imputation for time series.
\newblock In {\em Advances in Neural Information Processing Systems}, pages
  6776--6786, 2018.

\bibitem{62}
Han{-}Gyu Kim, Gil{-}Jin Jang, Ho{-}Jin Choi, Minho Kim, Young{-}Won Kim, and
  Jaehun Choi.
\newblock Recurrent neural networks with missing information imputation for
  medical examination data prediction.
\newblock In {\em {IEEE} International Conference on Big Data and Smart
  Computing}, pages 317--323, 2017.

\bibitem{54}
Qingxiong Tan, Mang Ye, Baoyao Yang, Siqi Liu, Andy~Jinhua Ma,
  Terry~Cheuk{-}Fung Yip, Grace~Lai{-}Hung Wong, and PongChi Yuen.
\newblock {DATA-GRU:} dual-attention time-aware gated recurrent unit for
  irregular multivariate time series.
\newblock In {\em The Thirty-Fourth {AAAI} Conference on Artificial
  Intelligence}, pages 930--937, 2020.

\bibitem{63}
Mohammad~Taha Bahadori and Zachary~Chase Lipton.
\newblock Temporal-clustering invariance in irregular healthcare time series.
\newblock {\em CoRR}, abs/1904.12206, 2019.

\bibitem{55}
Satya~Narayan Shukla and Benjamin~M. Marlin.
\newblock Interpolation-prediction networks for irregularly sampled time
  series.
\newblock In {\em International Conference on Learning Representations}, 2019.

\bibitem{56}
Zhengping Che, Sanjay Purushotham, Max~Guangyu Li, Bo~Jiang, and Yan Liu.
\newblock Hierarchical deep generative models for multi-rate multivariate time
  series.
\newblock In {\em Proceedings of the 35th International Conference on Machine
  Learning}, volume~80, pages 783--792, 2018.

\bibitem{32}
A.~et~al. Johnson.
\newblock Mimic-iii, a freely accessible critical care database.
\newblock {\em SCI. data}, 2016.

\bibitem{91}
Reyna~Matthew A., Josef~Christopher S, Jeter Russell, Shashikumar~Supreeth P.,
  Westover~M. Brandon, Nemati Shamim, Clifford~Gari D., and Sharma Ashish.
\newblock Early prediction of sepsis from clinical data: The
  physionet/computing in cardiology challenge 2019.
\newblock {\em Critical Care Medicine}, 48(121):210--217, 2020.

\bibitem{107}
Samaneh~Layeghian Javan, Mohammad~Mehdi Sepehri, Malihe~Layeghian Javan, and
  Toktam Khatibi.
\newblock An intelligent warning model for early prediction of cardiac arrest
  in sepsis patients.
\newblock {\em Comput. Methods Programs Biomed.}, 178:47--58, 2019.

\bibitem{108}
Yuxi Zhou, Shenda Hong, Junyuan Shang, Meng Wu, Qingyun Wang, Hongyan Li, and
  Junqing Xie.
\newblock K-margin-based residual-convolution-recurrent neural network for
  atrial fibrillation detection.
\newblock In {\em Proceedings of the Twenty-Eighth International Joint
  Conference on Artificial Intelligence}, pages 6057--6063, 2019.

\bibitem{109}
Shenda Hong, Yuxi Zhou, Junyuan Shang, Cao Xiao, and Jimeng Sun.
\newblock Opportunities and challenges of deep learning methods for
  electrocardiogram data: {A} systematic review.
\newblock {\em Comput. Biol. Medicine}, 122:103801, 2020.

\bibitem{110}
Adler~J. Perotte, Rajesh Ranganath, Jamie~S. Hirsch, David~M. Blei, and
  No{\'{e}}mie Elhadad.
\newblock Risk prediction for chronic kidney disease progression using
  heterogeneous electronic health record data and time series analysis.
\newblock {\em J. Am. Medical Informatics Assoc.}, 22(4):872--880, 2015.

\bibitem{46}
Steven~Cheng{-}Xian Li and Benjamin~M. Marlin.
\newblock Learning from irregularly-sampled time series: {A} missing data
  perspective.
\newblock {\em CoRR}, abs/2008.07599, 2020.

\bibitem{48}
G.~E.~P. Box and G.~M. Jenkins.
\newblock Time series analysis : forecasting and control.
\newblock {\em Journal of Time}, 31(3), 2010.

\bibitem{49}
Shivam Srivastava, Prithviraj Sen, and Berthold Reinwald.
\newblock Forecasting in multivariate irregularly sampled time series with
  missing values.
\newblock {\em CoRR}, abs/2004.03398, 2020.

\bibitem{26}
Yoon Jinsung, Zame~William R., and Mihaelaå Van~Der Schaar.
\newblock Estimating missing data in temporal data streams using
  multi-directional recurrent neural networks.
\newblock {\em IEEE Transactions on Biomedical Engineering}, PP:1--1, 2017.

\bibitem{28}
S~Hochreiter and J~Schmidhuber.
\newblock Long short-term memory.
\newblock {\em Neural Computation}, 9(8):1735--1780, 1997.

\bibitem{51}
Junyoung Chung, Caglar Gulcehre, Kyung~Hyun Cho, and Yoshua Bengio.
\newblock Empirical evaluation of gated recurrent neural networks on sequence
  modeling.
\newblock {\em Eprint Arxiv}, 2014.

\bibitem{53}
Yeo{-}Jin Kim and Min Chi.
\newblock Temporal belief memory: Imputing missing data during {RNN} training.
\newblock In {\em Proceedings of the Twenty-Seventh International Joint
  Conference on Artificial Intelligence}, pages 2326--2332, 2018.

\bibitem{44}
Matthew Herland, Taghi~M. Khoshgoftaar, and Randall Wald.
\newblock Survey of clinical data mining applications on big data in health
  informatics.
\newblock In {\em International Conference on Machine Learning and
  Applications, {ICMLA}}, pages 465--472, 2013.

\bibitem{45}
Suzan Arslanturk, Mohammad~Reza Siadat, Theophilus Ogunyemi, Kim Killinger, and
  Ananias Diokno.
\newblock Analysis of incomplete and inconsistent clinical survey data.
\newblock {\em Knowledge and Information Systems}, 46(3):731--750, 2016.

\bibitem{47}
Satya~Narayan Shukla and Benjamin~M. Marlin.
\newblock Modeling irregularly sampled clinical time series.
\newblock {\em CoRR}, abs/1812.00531, 2018.

\bibitem{13}
Yonghong Luo, Xiangrui Cai, Ying Zhang, Jun Xu, and Xiaojie Yuan.
\newblock Multivariate time series imputation with generative adversarial
  networks.
\newblock In {\em Advances in Neural Information Processing Systems}, pages
  1603--1614, 2018.

\bibitem{92}
A.~Goldberger, L.~Amaral, L.~Glass, J.~Hausdorff, P.~C. Ivanov, R.~Mark, and
  H.~E. Stanley.
\newblock Physiobank, physiotoolkit, and physionet: Components of a new
  research resource for complex physiologic signals.
\newblock {\em Circulation [Online]}, 101(23):pp. e215–e220, 2000.

\bibitem{58}
C.~Reyna, M.and~Josef, B.and Westover M.~B. Jeter, R.and Shashikumar
  S.and~Moody, Sharma A., Nemati S., and G.~Clifford.
\newblock Early prediction of sepsis from clinical data -- the physionet
  computing in cardiology challenge 2019 (version 1.0.0).
\newblock {\em PhysioNet}, 2019.

\bibitem{57}
Goncalves J et~al. Yan~L, Zhang H~T.
\newblock An interpretable mortality prediction model for covid-19 patients.
\newblock {\em Nature, Machine intelligence}, 2, 2020.

\bibitem{72}
Qingxiong Tan, Andy~Jinhua Ma, Mang Ye, Baoyao Yang, Huiqi Deng,
  Vincent~Wai{-}Sun Wong, Yee{-}Kit Tse, Terry~Cheuk{-}Fung Yip,
  Grace~Lai{-}Hung Wong, Jessica~Yuet{-}Ling Ching, Francis~Ka{-}Leung Chan,
  and Pong~C. Yuen.
\newblock {UA-CRNN:} uncertainty-aware convolutional recurrent neural network
  for mortality risk prediction.
\newblock In {\em Proceedings of the 28th {ACM} International Conference on
  Information and Knowledge Management}, pages 109--118, 2019.

\bibitem{73}
Qingxiong Tan, Andy~Jinhua Ma, Huiqi Deng, Vincent~Wai{-}Sun Wong, Yee{-}Kit
  Tse, Terry~Cheuk{-}Fung Yip, Grace~Lai{-}Hung Wong, Ching Ling~Jessica Yuet,
  Francis~Ka{-}Leung Chan, and Pong~C. Yuen.
\newblock A hybrid residual network and long short-term memory method for
  peptic ulcer bleeding mortality prediction.
\newblock In {\em {AMIA} 2018, American Medical Informatics Association Annual
  Symposium}, 2018.

\bibitem{74}
Yanbo Xu, Siddharth Biswal, Shriprasad~R. Deshpande, Kevin~O. Maher, and Jimeng
  Sun.
\newblock {RAIM:} recurrent attentive and intensive model of multimodal patient
  monitoring data.
\newblock In Yike Guo and Faisal Farooq, editors, {\em Proceedings of the 24th
  {ACM} {SIGKDD} International Conference on Knowledge Discovery and Data
  Mining}, pages 2565--2573, 2018.

\bibitem{6}
Buckley Jonathan and James Ian.
\newblock Linear regression with censored data.
\newblock {\em Biometrika}, (3):429--436, 1979.

\bibitem{7}
R~Williams and D~Zipser.
\newblock A learning algorithm for continually running fully recurrent neural
  networks.
\newblock {\em Neural Computation}, 1(2):270--280, 2014.

\bibitem{15}
Martin~W. Dünser, Jukka Takala, Hanno Ulmer, Viktoria~D. Mayr, Günter
  Luckner, Stefan Jochberger, Fritz Daudel, Philipp Lepper, Walter~R.
  Hasibeder, and Stephan~M. Jakob.
\newblock Arterial blood pressure during early sepsis and outcome.
\newblock {\em Intensive Care Medicine}, 35(7):1225--1233, 2009.

\bibitem{14}
Vincent Liu, Gabriel~J. Escobar, John~D. Greene, Jay Soule, Alan Whippy,
  Derek~C. Angus, and Theodore~J. Iwashyna.
\newblock Hospital deaths in patients with sepsis from 2 independent cohorts.
\newblock {\em Jama}, 312(1):90--92, 2014.

\bibitem{16}
Xu~Chu, Ihab~F. Ilyas, Sanjay Krishnan, and Jiannan Wang.
\newblock Data cleaning: Overview and emerging challenges.
\newblock In {\em International Conference on Management of Data}, 2016.

\bibitem{19}
David~M. Kreindler and Charles~J. Lumsden.
\newblock The effects of the irregular sample and missing data in time series
  analysis.
\newblock {\em Nonlinear Dynamics Psychology and Life ences}, 10(2):187--214,
  2006.

\bibitem{20}
Todd Ogden.
\newblock Wavelet methods for time series analysis. (book reviews).
\newblock {\em Journal of the American}, (March), 2002.

\bibitem{21}
Rehfeld, K., Marwan, N., Heitzig, J., Kurths, and J.
\newblock Comparison of correlation analysis techniques for irregularly sampled
  time series.
\newblock {\em Nonlinear Processes in Geophysics}, 2011.

\bibitem{22}
Patrick~Royston Ian R~White and Angela~M Wood.
\newblock Multiple imputation using chained equations. issues and guidance for
  practice.
\newblock {\em Statistics in medicine}, 30(2):377–399, 2011.

\bibitem{23}
Pedro~J. García-Laencina, José-Luis Sancho-Gómez, and Aníbal~R.
  Figueiras-Vidal.
\newblock Pattern classification with missing data: a review.
\newblock {\em Neural Computing and Applications}, 19(2):263--282, 2010.

\bibitem{75}
Volker Tresp and Thomas Briegel.
\newblock A solution for missing data in recurrent neural networks with an
  application to blood glucose prediction.
\newblock In {\em Advances in Neural Information Processing Systems}, pages
  971--977, 1997.

\bibitem{76}
Shahla Parveen and Phil~D. Green.
\newblock Speech recognition with missing data using recurrent neural nets.
\newblock In {\em Advances in Neural Information Processing Systems}, pages
  1189--1195, 2001.

\bibitem{50}
Alex Graves and Jürgen Schmidhuber.
\newblock Framewise phoneme classification with bidirectional lstm and other
  neural network architectures.
\newblock {\em Neural Networks}, 18(5–6):602--610, 2005.

\bibitem{59}
Jaeyoon Kim, Donghyun Tae, and Junhee Seok.
\newblock A survey of missing data imputation using generative adversarial
  networks.
\newblock In {\em 2020 International Conference on Artificial Intelligence in
  Information and Communication}, pages 454--456, 2020.

\bibitem{77}
Guanghao Zhang, Enmei Tu, and Dongshun Cui.
\newblock Stable and improved generative adversarial nets {(GANS):} {A}
  constructive survey.
\newblock In {\em 2017 {IEEE} International Conference on Image Processing},
  pages 1871--1875, 2017.

\bibitem{60}
Jinsung Yoon, James Jordon, and Mihaela van~der Schaar.
\newblock {GAIN:} missing data imputation using generative adversarial nets.
\newblock In {\em Proceedings of the 35th International Conference on Machine
  Learning}, volume~80, pages 5675--5684, 2018.

\bibitem{78}
Ramiro~Daniel Camino, Christian~A. Hammerschmidt, and Radu State.
\newblock Improving missing data imputation with deep generative models.
\newblock {\em CoRR}, abs/1902.10666, 2019.

\bibitem{79}
H~Zhang and DP~Woodruff.
\newblock Medical missing data imputation by stackelberg gan.
\newblock {\em Carnegie Mellon University}, 2018.

\bibitem{25}
Amy S~Nowacki Brian J~Wells, Kevin M~Chagin and Michael~W Kattan.
\newblock Strategies for handling missing data in electronic health record
  derived data.
\newblock {\em Generating Evidence and Methods}, 1(3), 2010.

\bibitem{83}
Simon Haykin.
\newblock {\em Kalman Filtering and Neural Networks}.
\newblock John Wiley and Sons, Inc, 2001.

\bibitem{84}
Pederson and P.~Shane.
\newblock Hidden markov and other models for discretevalued time serie.
\newblock {\em Technometrics}, 40(3):263--263, 1998.

\bibitem{81}
Mikolaj Binkowski, Gautier Marti, and Philippe Donnat.
\newblock Autoregressive convolutional neural networks for asynchronous time
  series.
\newblock In {\em Proceedings of the 35th International Conference on Machine
  Learning}, volume~80, pages 579--588, 2018.

\bibitem{80}
Steven~Cheng{-}Xian Li and Benjamin~M. Marlin.
\newblock A scalable end-to-end gaussian process adapter for irregularly
  sampled time series classification.
\newblock In {\em Advances in Neural Information Processing Systems}, pages
  1804--1812, 2016.

\bibitem{82}
Joseph Futoma, Sanjay Hariharan, and Katherine~A. Heller.
\newblock Learning to detect sepsis with a multitask gaussian process {RNN}
  classifier.
\newblock In {\em Proceedings of the 34th International Conference on Machine
  Learning}, volume~70, pages 1174--1182, 2017.

\bibitem{65}
Edward Choi, Mohammad~Taha Bahadori, Andy Schuetz, Walter~F. Stewart, and
  Jimeng Sun.
\newblock Doctor {AI:} predicting clinical events via recurrent neural
  networks.
\newblock In {\em Proceedings of the 1st Machine Learning in Health Care},
  volume~56, pages 301--318, 2016.

\bibitem{68}
Yijun Li, Sifei Liu, Jimei Yang, and Ming{-}Hsuan Yang.
\newblock Generative face completion.
\newblock In {\em 2017 {IEEE} Conference on Computer Vision and Pattern
  Recognition}, pages 5892--5900, 2017.

\bibitem{66}
Ian Goodfellow, Jean Pouget-Abadie, Mehdi Mirza, Bing Xu, and David
  Warde-Farley.
\newblock Generative adversarial nets.
\newblock In {\em Advances in neural information processing systems}, page
  2672–2680, 2014.

\bibitem{67}
Ashish Bora, Eric Price, and Alexandros~G. Dimakis.
\newblock Ambientgan: Generative models from lossy measurements.
\newblock In {\em 6th International Conference on Learning Representations},
  2018.

\bibitem{69}
Shuang Liu, Olivier Bousquet, and Kamalika Chaudhuri.
\newblock Approximation and convergence properties of generative adversarial
  learning.
\newblock In {\em Advances in Neural Information Processing Systems}, pages
  5545--5553, 2017.

\bibitem{70}
Lantao Yu, Weinan Zhang, Jun Wang, and Yong Yu.
\newblock Seqgan: Sequence generative adversarial nets with policy gradient.
\newblock In {\em Proceedings of the Thirty-First {AAAI} Conference on
  Artificial Intelligence}, pages 2852--2858, 2017.

\bibitem{61}
Steven~Cheng{-}Xian Li, Bo~Jiang, and Benjamin~M. Marlin.
\newblock Misgan: Learning from incomplete data with generative adversarial
  networks.
\newblock In {\em 7th International Conference on Learning Representations},
  2019.

\bibitem{85}
Diederik~P. Kingma and Jimmy Ba.
\newblock Adam: {A} method for stochastic optimization.
\newblock In {\em 3rd International Conference on Learning Representations},
  2015.

\bibitem{86}
Imane Ezzine and Laila Benhlima.
\newblock A study of handling missing data methods for big data.
\newblock In {\em 2018 IEEE 5th International Congress on Information Science
  and Technology (CiSt)}, 2018.

\bibitem{87}
Benjamin King and D.~B. Rubin.
\newblock Multiple imputation for nonresponse in surveys.
\newblock {\em Journal of the American Statal Association}, 84(406):612, 1988.

\bibitem{88}
Rahul Mazumder, Trevor Hastie, and Robert Tibshirani.
\newblock Spectral regularization algorithms for learning large incomplete
  matrices.
\newblock {\em Journal of Machine Learning Research Jmlr}, 11(11):2287, 2009.

\bibitem{89}
Hsiang{-}Fu Yu, Nikhil Rao, and Inderjit~S. Dhillon.
\newblock Temporal regularized matrix factorization for high-dimensional time
  series prediction.
\newblock In {\em Advances in Neural Information Processing Systems}, pages
  847--855, 2016.

\bibitem{90}
Christoph Molnar.
\newblock {\em Interpretable Machine Learning: A Guide for Making Black Box
  Models Explainable}.
\newblock online, 2020.

\bibitem{111}
Gregor Stiglic, Primoz Kocbek, Nino Fijacko, Marinka Zitnik, Katrien Verbert,
  and Leona Cilar.
\newblock Interpretability of machine learning-based prediction models in
  healthcare.
\newblock {\em Wiley Interdiscip. Rev. Data Min. Knowl. Discov.}, 10(5), 2020.

\bibitem{120}
Nikos Fazakis, Georgios Kostopoulos, Sotiris Kotsiantis, and Iosif Mporas.
\newblock Iterative robust semi-supervised missing data imputation.
\newblock {\em {IEEE} Access}, 8:90555--90569, 2020.

\bibitem{122}
Saloni Dash, Andrew Yale, Isabelle Guyon, and Kristin~P. Bennett.
\newblock Medical time-series data generation using generative adversarial
  networks.
\newblock In {\em Artificial Intelligence in Medicine}, volume 12299, pages
  382--391, 2020.

\bibitem{113}
Oguzhan Karaahmetoglu, Fatih Ilhan, Ismail Balaban, and Suleyman~Serdar Kozat.
\newblock Unsupervised online anomaly detection on irregularly sampled or
  missing valued time-series data using {LSTM} networks.
\newblock {\em CoRR}, abs/2005.12005, 2020.

\bibitem{112}
Ahmed Guecioueur and Franz~J. Kir{\'{a}}ly.
\newblock Kernels for time series with irregularly-spaced multivariate
  observations.
\newblock {\em CoRR}, abs/2004.08545, 2020.

\bibitem{114}
Yang Jiao, Kai Yang, Shaoyu Dou, Pan Luo, Sijia Liu, and Dongjin Song.
\newblock Timeautoml: Autonomous representation learning for multivariate
  irregularly sampled time series.
\newblock {\em CoRR}, abs/2010.01596, 2020.

\bibitem{117}
Zhenning Mei, Xian Zhao, and Hongyu Chen.
\newblock A distributed descriptor characterizing structural irregularity of
  {EEG} time series for epileptic seizure detection.
\newblock In {\em 40th Annual International Conference of the {IEEE}
  Engineering in Medicine and Biology Society}, pages 3386--3389, 2018.

\bibitem{118}
Y.~Nancy Jane, Harichandran~Khanna Nehemiah, and Arputharaj Kannan.
\newblock A bio-statistical mining approach for classifying multivariate
  clinical time series data observed at irregular intervals.
\newblock {\em Expert Syst. Appl.}, 78:283--300, 2017.

\bibitem{119}
Santosh Tirunagari, Simon~C. Bull, and Norman Poh.
\newblock Automatic classification of irregularly sampled time series with
  unequal lengths: {A} case study on estimated glomerular filtration rate.
\newblock In {\em 26th {IEEE} International Workshop on Machine Learning for
  Signal Processing}, pages 1--6, 2016.

\bibitem{123}
Lutong Wang, Hong Wang, Yongqiang Song, and Qian Wang.
\newblock Mcpl-based {FT-LSTM:} medical representation learning-based clinical
  prediction model for time series events.
\newblock {\em {IEEE} Access}, 7:70253--70264, 2019.

\bibitem{115}
Manxia Liu, Fabio Stella, Arjen Hommersom, Peter J.~F. Lucas, Lonneke Boer, and
  Erik Bischoff.
\newblock A comparison between discrete and continuous time bayesian networks
  in learning from clinical time series data with irregularity.
\newblock {\em Artif. Intell. Medicine}, 95:104--117, 2019.

\bibitem{116}
Bhanu~Pratap Singh, Iman Deznabi, Bharath Narasimhan, Bryon Kucharski, Rheeya
  Uppaal, Akhila Josyula, and Madalina Fiterau.
\newblock Multi-resolution networks for flexible irregular time series modeling
  (multi-fit).
\newblock {\em CoRR}, abs/1905.00125, 2019.

\end{thebibliography}

\end{document}